\documentclass{article}

 \usepackage[main, final]{neurips_2026}

\usepackage[utf8]{inputenc} 
\usepackage[T1]{fontenc}    
\usepackage{hyperref}       
\usepackage{url}            
\usepackage{booktabs}       
\usepackage{amsfonts}       
\usepackage{nicefrac}       
\usepackage{microtype}      
\usepackage{xcolor}         
\usepackage{array,multirow,graphicx}
\usepackage{amsmath}
\usepackage{mathtools}
\usepackage{bbm}
\usepackage{hhline}
\usepackage{boldline}
\usepackage{tabularx}
\usepackage{float}
\usepackage{comment}
\usepackage{color}
\usepackage{amssymb}
\usepackage{multicol}
\usepackage{algorithm}
\usepackage{algpseudocode}
\usepackage{soul} 
\usepackage{subcaption}
\usepackage{wrapfig}
\usepackage{lipsum}
\usepackage{booktabs}
\usepackage{colortbl}
\usepackage{nicematrix}
\usepackage{kotex}
\usepackage{placeins}
\usepackage{pifont}
\usepackage{caption}
\usepackage{algorithm}
\usepackage{algpseudocode}
\usepackage{placeins}
\usepackage{makecell}

\title{TF-PRVR: Training-Free Partially Relevant \\ Video Retrieval}

\author{
    Giyeol Kim \quad
    Chanho Eom\textsuperscript{\textdagger} \\
    Chung-Ang University \\
    \texttt{\{giyeolkim, cheom\}@cau.ac.kr} \\
    \href{https://perceptualai-lab.github.io/TF-PRVR/}
    {\texttt{https://perceptualai-lab.github.io/TF-PRVR/}}
}

\begin{document}

\maketitle

\begingroup
\renewcommand{\thefootnote}{\fnsymbol{footnote}}
\footnotetext[2]{Corresponding author.}
\endgroup

\begin{abstract}
Partially Relevant Video Retrieval (PRVR) aims to retrieve untrimmed videos containing moments relevant to a given text query. Despite recent progress, existing PRVR methods suffer from two key limitations: a fixed video decomposition scheme that causes semantic dilution, and source-domain overfitting induced by task-specific training. In this paper, we propose TF-PRVR, the first training-free framework for PRVR. TF-PRVR leverages frozen vision-language features to construct video-specific hierarchical representations. It derives temporal semantic signals from frame-level features and applies frequency-based multi-scale analysis to identify adaptive temporal boundaries, producing hierarchical segments with coherent event-level semantics. Built on these segments, TF-PRVR constructs a unified multi-scale graph and propagates query relevance across temporally and semantically related nodes. A moment-aware scoring strategy then aggregates temporally aligned relevance across scales, emphasizing consistently supported moments while suppressing isolated false responses. Without task-specific training, TF-PRVR preserves the general-purpose alignment capability of pre-trained vision-language models and avoids dataset-specific overfitting. Extensive experiments demonstrate consistent performance across datasets with diverse visual and temporal characteristics, suggesting a practical direction for training-free PRVR.
\end{abstract}

\section{Introduction}
\label{sec:intro}
With the rapid growth of social media and video-sharing platforms, the demand for retrieval systems that can identify desired video content from large-scale video corpora has grown substantially. Text-to-Video Retrieval (T2VR)~\cite{jin2022expectation, jin2023video, jin2023text, jin2023diffusionret, ma2022x} addresses this need by mapping text and video representations into a shared embedding space, thereby enabling users to search videos with natural language queries. However, conventional T2VR methods operate under the implicit assumption that each video is pre-trimmed to be fully relevant to its corresponding text query. This assumption may not hold in practice, as real-world videos are often untrimmed and contain multiple distinct events.

\begin{figure}
    \centering
    \includegraphics[width=1\linewidth]{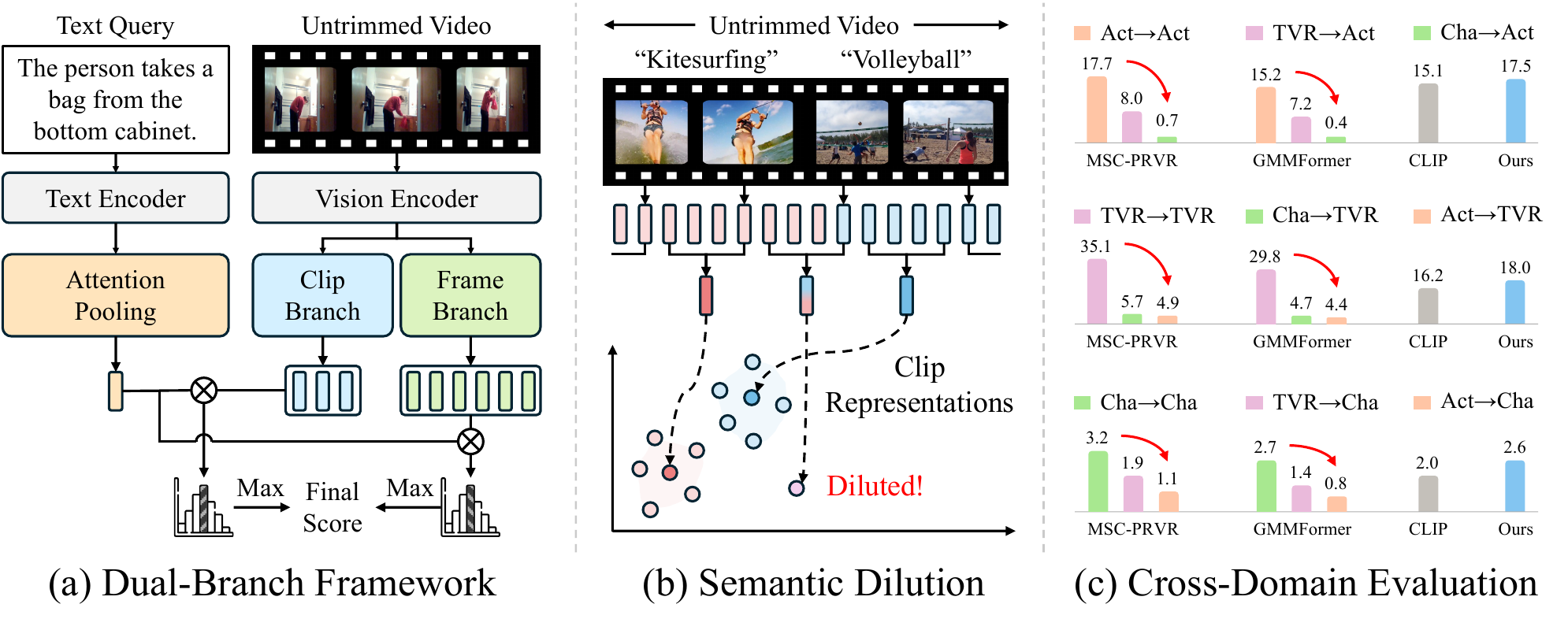}
    \vspace{-1.8em}
    \caption{
(a) Dual-branch framework of existing PRVR methods~\cite{ms-sl, msc-prvr, gmmformer, hlformer, proto}.
(b) Semantic dilution caused by fixed-size clip aggregation.
(c) Cross-dataset comparison (R@1 (\%)) on three datasets (TVR~\cite{tvr}, ActivityNet Captions (Act)~\cite{activitynet}, Charades-STA (Cha)~\cite{charades-sta}) using CLIP-L~\cite{CLIP} features. Each group shows in-domain (\textit{e.g.}, Act$\rightarrow$Act) and cross-domain (\textit{e.g.}, TVR$\rightarrow$Act) results for existing methods~\cite{gmmformer, msc-prvr}, zero-shot CLIP, and our training-free TF-PRVR.
(Best viewed in color.)
}
\vspace{-0.8em}
    \label{fig:teaser}
\end{figure}

To bridge this gap, partially relevant video retrieval (PRVR)~\cite{ms-sl, gmmformer, proto, hlformer} has emerged as a more practical setting, aiming to retrieve untrimmed videos where only specific video segments correspond to the given query. Since annotating the precise temporal boundaries of query-relevant moments in untrimmed videos is labor-intensive, existing PRVR methods do not rely on moment-level supervision. Instead, as shown in Fig.~\ref{fig:teaser} (a), they typically leverage pre-trained vision-language models~\cite{CLIP} to extract visual and textual features, and encode videos with two complementary branches: a frame branch for capturing fine-grained local semantics and a clip branch for modeling broader temporal contexts. Based on this dual-branch framework, prior methods have mainly focused on how to effectively model clip representations, such as explicit sliding-window construction~\cite{ms-sl}, Gaussian-constrained implicit modeling~\cite{gmmformer, gmmformerv2, hlformer}, and prototype-based compression~\cite{proto, sdm}.

Despite these advances, existing PRVR methods~\cite{ms-sl, gmmformer, gmmformerv2, hlformer, msc-prvr, proto, sdm} still face two critical challenges. The first challenge lies in the fixed video decomposition scheme that neglects the diverse temporal structures inherent in real-world videos. Videos naturally vary in their overall length, event density, and the duration of distinct events. Although the dual-branch framework (Fig.~\ref{fig:teaser} (a)) attempts to capture multi-granularity features, applying the same predefined two-level hierarchy to all videos may not be sufficient to reflect their diverse temporal characteristics. Moreover, in the clip branch, existing methods aggregate consecutive frames into a fixed number of clip representations to construct potential query-relevant moments. Such aggregation may cause semantic dilution (Fig.~\ref{fig:teaser} (b)), where frames from distinct events are merged into a single clip representation. For instance, when frames from the ``Kitesurfing'' and ``Volleyball'' events are pooled together, the resulting clip representation becomes semantically ambiguous, degrading the discriminability of query-relevant moments.

The second challenge arises from source-domain overfitting in existing training-based PRVR methods. As shown in Fig.~\ref{fig:teaser} (c), while prior works~\cite{gmmformer, msc-prvr} achieve promising performance under in-domain evaluation, their retrieval accuracy drops significantly when transferred to unseen target datasets. For instance, MSC-PRVR~\cite{msc-prvr} achieves an R@1 of 35.1\% when trained and evaluated on TVR~\cite{tvr}, but its performance drops to 5.7\% and 4.9\% when trained on Charades-STA~\cite{charades-sta} and ActivityNet Captions~\cite{activitynet}, respectively. Notably, these results fall substantially below those of the zero-shot CLIP~\cite{CLIP} baseline (16.2\%). These results suggest that, despite leveraging the rich semantic knowledge of pre-trained vision-language models~\cite{CLIP}, the trainable modules tend to adapt to source-specific statistical patterns, leading to degraded retrieval performance on unseen target datasets. Such sensitivity to the source training distribution limits their applicability to heterogeneous video collections.

In this paper, we propose TF-PRVR, the first training-free framework for partially relevant video retrieval. We first revisit existing PRVR methods~\cite{ms-sl, gmmformer, gmmformerv2, hlformer, msc-prvr, proto, sdm} through an in-depth analysis and identify two critical limitations: source-domain overfitting induced by task-specific training and fixed video decomposition that overlooks video-specific temporal structures. Motivated by these findings, TF-PRVR avoids PRVR-specific optimization and instead leverages the general semantic alignment of frozen vision-language models.
Specifically, it derives temporal semantic signals from frame-level features and performs frequency-based multi-scale analysis to detect adaptive temporal boundaries, constructing hierarchical segments that preserve coherent event-level semantics and reduce semantic dilution. TF-PRVR then builds a query-independent unified multi-scale graph over these segments, over which query relevance is propagated to enforce temporal and semantic consistency. Finally, a moment-aware scoring strategy aggregates temporally aligned relevance across scales, emphasizing consistently supported moments while suppressing isolated responses. Extensive experiments demonstrate consistent training-free retrieval performance across datasets with diverse visual and temporal characteristics, highlighting the effectiveness of our training-free design.

\section{Related Work}
\label{sec:related_work}
\vspace{-0.3em}
\noindent\textbf{Partially Relevant Video Retrieval.} 
Conventional text-to-video retrieval methods~\cite{jin2022expectation, jin2023video, jin2023text, jin2023diffusionret, wang2022hybrid, ma2022x} assume that videos are pre-trimmed and fully relevant to their paired queries, which limits their applicability to real-world untrimmed videos. To address this, partially relevant video retrieval (PRVR)~\cite{ms-sl, gmmformer, proto, arl, msc-prvr, hlformer, sdm, dl-dkd} has been introduced as a more practical setting, where a video is considered relevant if it contains at least one moment corresponding to the given query. MS-SL~\cite{ms-sl} first formulates PRVR as a multiple instance learning problem and proposes a dual-branch framework that encodes videos at both frame and clip scales via multi-scale sliding windows. However, exhaustive clip construction introduces substantial redundancy and storage overhead. To alleviate this, GMMFormer~\cite{gmmformer} proposes implicit clip modeling by incorporating Gaussian Mixture Model constraints into frame-level self-attention.  ProtoPRVR~\cite{proto} compresses diverse video contexts into compact prototypes to balance retrieval effectiveness and efficiency, and HLFormer~\cite{hlformer} introduces hyperbolic space learning to better capture the hierarchical structure of untrimmed videos. Despite these advances, existing methods still rely on fixed video decomposition and task-specific training, making them vulnerable to semantic dilution and domain overfitting. In contrast, we address these issues with video-specific hierarchical representations and a fully training-free retrieval pipeline.

\noindent\textbf{Frequency-Based Temporal Modeling.}
Frequency-domain analysis has been widely used to reveal temporal structures that are difficult to capture in the raw time domain. Early studies exploit Fourier-based representations for action recognition and video captioning, such as HPM~\cite{rahmani20163d}, which introduces a group-sparse Fourier Temporal Pyramid. Wavelet-based methods have also been used to model non-stationary temporal patterns, as in QUVA Repetition~\cite{runia2018real}, which applies continuous wavelet transforms to optical-flow signals for repetition estimation. More recently, STMFANet~\cite{jin2020exploring} applies discrete wavelet transforms along spatial and temporal axes for multi-frequency video prediction. Inspired by these works, we first introduce frequency-based temporal modeling to PRVR. Unlike prior methods that mainly analyze low-level motion or pixel-domain signals, we decompose semantic transitions in frozen vision-language features to identify video-specific temporal boundaries, enabling adaptive hierarchical representation without task-specific training or moment-level annotations.

\begin{figure}[t]
\centering

\begin{minipage}[t]{0.40\textwidth}
\vspace{0pt}
\centering

\includegraphics[width=0.9\linewidth]{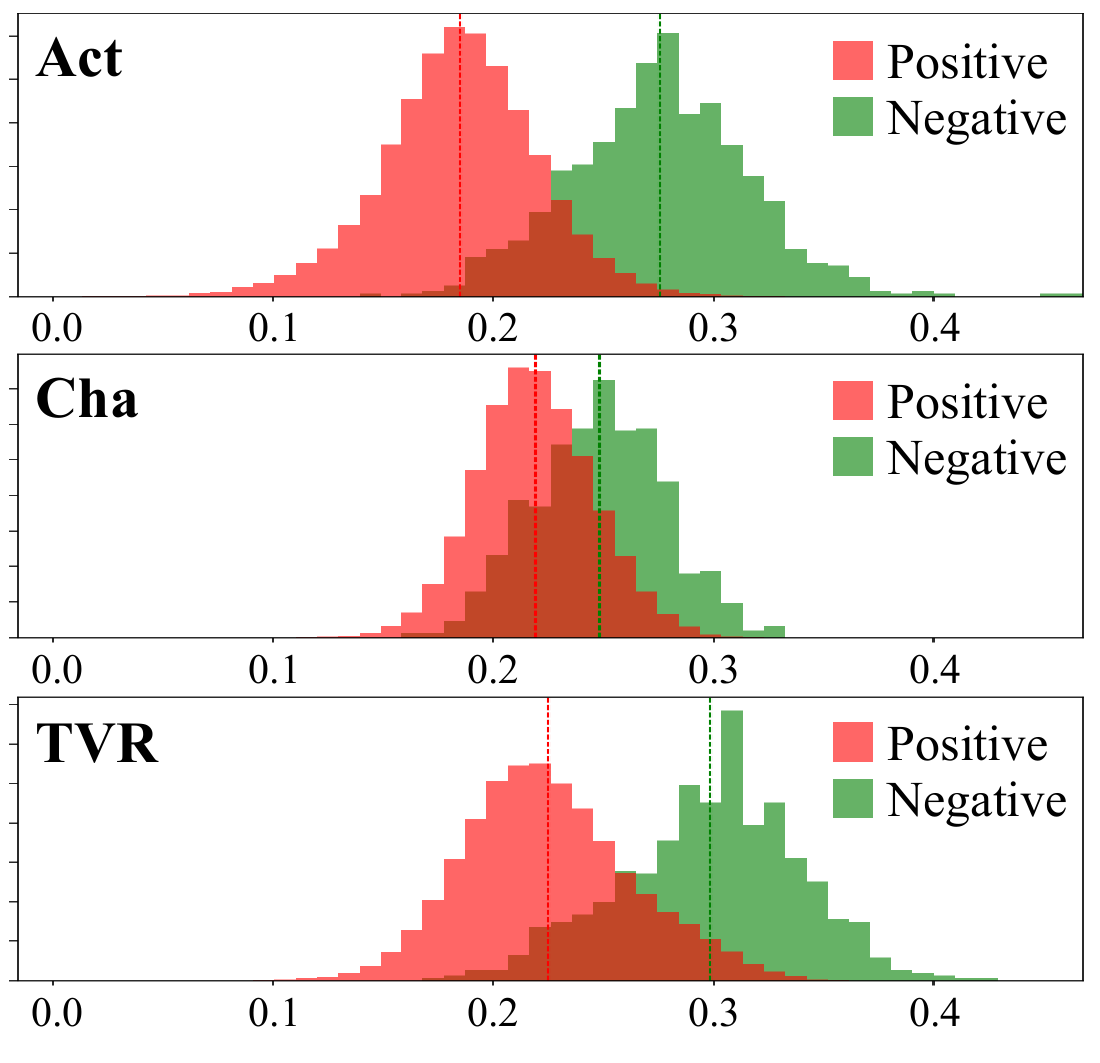}
\vspace{-0.8em}
\caption{Cosine similarity distributions of positive and negative text-video pairs.}
\label{fig:sim}
\end{minipage}%
\hfill
\begin{minipage}[t]{0.57\textwidth}
\vspace{0pt}
\centering

\captionof{table}{Maximum Mean Discrepancy (MMD) comparison of feature distributions across dataset pairs. MMD is computed on query-, clip-, and frame-level features between each dataset pair. Lower MMD indicates smaller cross-dataset distribution discrepancy.}
\vspace{-0.2em}
\label{tab:mmd}

{
\renewcommand{\arraystretch}{1.0}
\resizebox{\linewidth}{!}{
\begin{tabular}{llccc}
\toprule
Feature & Model & TVR vs Cha & TVR vs Act & Cha vs Act \\
\midrule
\multirow{3}{*}{Query}
& CLIP-L~\cite{CLIP}    & 0.002 & 0.001 & 0.002 \\
& HLFormer~\cite{hlformer}  & 0.054 & 0.013 & 0.051 \\
& MSC-PRVR~\cite{msc-prvr}  & 0.063 & 0.019 & 0.048 \\
\midrule
\multirow{3}{*}{Clip}
& CLIP-L~\cite{CLIP}    & 0.019 & 0.017 & 0.003 \\
& HLFormer~\cite{hlformer}  & 0.031 & 0.029 & 0.014 \\
& MSC-PRVR~\cite{msc-prvr}  & 0.030 & 0.031 & 0.016 \\
\midrule
\multirow{3}{*}{Frame}
& CLIP-L~\cite{CLIP}    & 0.011 & 0.014 & 0.002 \\
& HLFormer~\cite{hlformer}  & 0.030 & 0.025 & 0.013 \\
& MSC-PRVR~\cite{msc-prvr}  & 0.029 & 0.028 & 0.018 \\
\bottomrule
\end{tabular}
}
}

\end{minipage}
\end{figure}

\section{Revisiting the Limitations of Existing PRVR Methods}
\label{sec:analysis}

In this section, we empirically analyze two limitations of existing PRVR methods: source-domain overfitting from task-specific training (Sec.~\ref{sec:domain}) and rigid video decomposition (Sec.~\ref{sec:scheme}).

\subsection{Analysis of Source-Domain Overfitting}
\label{sec:domain}
To investigate source-domain overfitting in training-based PRVR methods, we first examine text-video similarity distributions. For each dataset, we compute cosine similarities between text queries and videos and plot the distributions of positive pairs and sampled negative pairs in Fig.~\ref{fig:sim}. The results show that both the score ranges and positive-negative margins differ across datasets. ActivityNet Captions~\cite{activitynet} shows a relatively clear separation, while Charades-STA~\cite{charades-sta} has a narrower margin with stronger overlap between positives and negatives. TVR~\cite{tvr} exhibits a distinct distribution pattern. This suggests that ranking objectives optimized on a single dataset may become sensitive to dataset-specific score distributions, limiting their transferability to unseen datasets. To further quantify this discrepancy, we measure Maximum Mean Discrepancy (MMD)~\cite{MMD} between feature distributions across dataset pairs. As shown in Table~\ref{tab:mmd}, frozen CLIP-L~\cite{CLIP} features maintain small cross-dataset MMD values (\textit{e.g.}, 0.002 between TVR and Cha), while trained PRVR models exhibit substantially larger values (\textit{e.g.}, 0.054 and 0.063 for HLFormer~\cite{hlformer} and MSC-PRVR~\cite{msc-prvr}). These results suggest that task-specific training amplifies cross-dataset distribution discrepancies relative to frozen CLIP features, consistent with performance degradation on unseen target datasets.

\subsection{Analysis of Fixed Video Decomposition}
\label{sec:scheme}

\begin{figure}[t]
\centering

\begin{minipage}[t]{0.40\textwidth}
\vspace{0pt}
\centering

\captionof{table}{Temporal span statistics under fixed-size clip sampling (32 clips).}
\vspace{-0.6em}
\label{tab:temporal_span}
\resizebox{\linewidth}{!}{
\begin{tabular}{lccc}
\toprule
Datasets & Avg. Span & Min. Span & Max. Span \\
\midrule
Act~\cite{activitynet}  & 204.01 & 2  & 1415 \\
Cha~\cite{charades-sta} & 91.39  & 16 & 606  \\
TVR~\cite{tvr}          & 50.87  & 2  & 181  \\
\bottomrule
\end{tabular}
}

\captionof{table}{SumR performance comparison of existing methods across video length groups on three benchmarks.}
\label{tab:video_length}
\resizebox{\linewidth}{!}{
\begin{tabular}{llccc}
\toprule
\multirow{2}{*}{Datasets} 
& \multirow{2}{*}{Model} 
& \multicolumn{3}{c}{Video Length} \\
\cmidrule(lr){3-5}
& & Full & Short & Long \\
\midrule
\multirow{3}{*}{Act~\cite{activitynet}}
& MS-SL~\cite{ms-sl}     & 140.1 & 122.3 & 148.3 \\
& GMMFormer~\cite{gmmformer} & 146.0 & 132.6 & 152.3 \\
& HLFormer~\cite{hlformer}  & 154.9 & 144.0 & 157.4 \\
\midrule
\multirow{3}{*}{Cha~\cite{charades-sta}}
& MS-SL~\cite{ms-sl}     & 68.4 & 67.6 & 65.3 \\
& GMMFormer~\cite{gmmformer} & 72.9 & 76.0 & 64.4 \\
& HLFormer~\cite{hlformer}  & 78.7 & 79.5 & 72.1 \\
\midrule
\multirow{3}{*}{TVR~\cite{tvr}}
& MS-SL~\cite{ms-sl}     & 172.4 & 181.2 & 167.2 \\
& GMMFormer~\cite{gmmformer} & 176.6 & 187.2 & 169.6 \\
& HLFormer~\cite{hlformer}  & 187.7 & 199.1 & 178.9 \\
\bottomrule
\end{tabular}
}

\captionof{table}{Dataset statistics including average moment and video lengths.}
\label{tab:dataset_stats}
\resizebox{\linewidth}{!}{
\begin{tabular}{lccccc}
\toprule
\multirow{2}{*}{Datasets} 
& \multicolumn{2}{c}{Average Length} 
& \multicolumn{3}{c}{Moment-to-Video Ratio} \\
\cmidrule(lr){2-3} \cmidrule(lr){4-6}
& Moments & Videos & Min & Max & Mean \\
\midrule
Act~\cite{activitynet} & 36.2s & 117.6s & 0.48\% & 100\% & 30.8\% \\
Cha~\cite{charades-sta} & 8.1s  & 30.0s  & 4.3\%  & 100\% & 26.3\% \\
TVR~\cite{tvr} & 9.1s  & 76.2s  & 0.48\% & 100\% & 11.9\% \\
\bottomrule
\end{tabular}
}
\end{minipage}
\hfill
\begin{minipage}[t]{0.58\textwidth}
\vspace{0pt}
\centering

\includegraphics[width=0.99\linewidth]{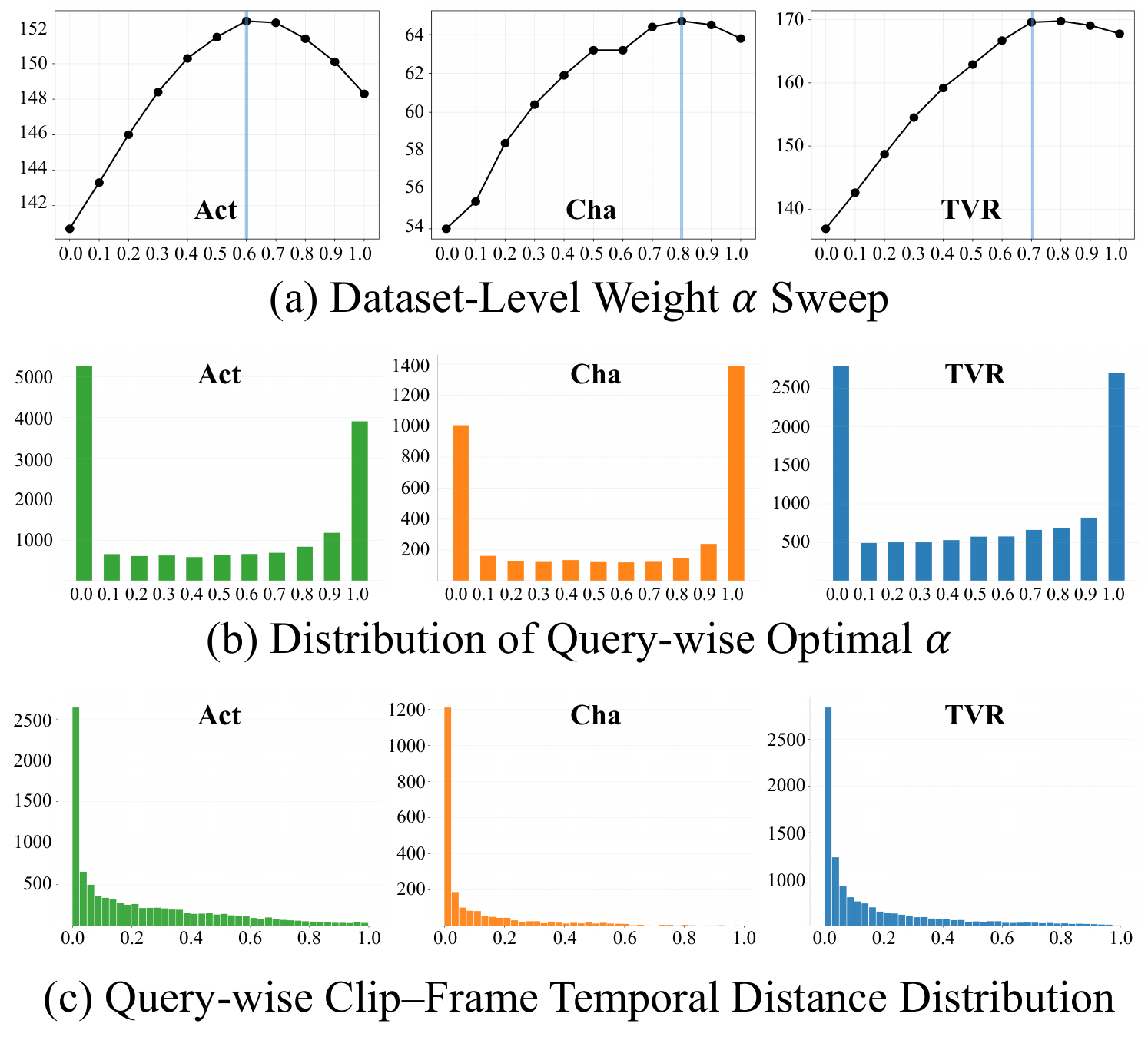}
\vspace{-0.7em}
\captionof{figure}{
(a) Retrieval performance (SumR) across datasets when varying the weighting factor $\alpha$.
(b) Distribution of query-wise optimal $\alpha$ values obtained by selecting the best-performing weight for each query in the test set.
(c) Distribution of normalized temporal distances between the clip and frame representations selected by max-pooling.
}
\label{fig:anlysis}

\end{minipage}
\end{figure}

\noindent\textbf{Semantic Dilution.}
Previous PRVR methods~\cite{ms-sl, gmmformer, gmmformerv2, hlformer, msc-prvr, proto} construct clip-level video representations by compressing frame features into a fixed number of clips (\textit{e.g.}, 32) via average pooling. As shown in Table~\ref{tab:temporal_span}, each clip spans approximately 204.1 frames on ActivityNet Captions (Act)~\cite{activitynet}, 91.4 on Charades-STA (Cha)~\cite{charades-sta}, and 50.9 on TVR~\cite{tvr} on average. Such content-agnostic aggregation can cause semantic dilution, where discriminative visual cues are mixed with unrelated events and become semantically ambiguous. To examine how fixed decomposition interacts with different temporal characteristics, as shown in Table~\ref{tab:video_length}, we split each dataset into short and long video groups. On ActivityNet Captions, where moments are long relative to the video (moment-to-video ratio of 30.8\% in Table~\ref{tab:dataset_stats}), short videos suffer since long moments are fragmented across multiple clips, while long videos benefit as each clip better aligns with the moment duration. As a result, existing methods consistently perform better on long videos (\textit{e.g.}, 157.4 vs. 144.0 for HLFormer~\cite{hlformer}, 152.3 vs. 132.6 for GMMFormer~\cite{gmmformer}). The opposite trend holds on TVR, where moments are short (moment-to-video ratio of 11.9\%), long videos suffer from semantic dilution as each clip aggregates unrelated events, while short videos better isolate the moment (\textit{e.g.}, 178.9 vs. 199.1 for HLFormer~\cite{hlformer}, 169.6 vs. 187.2 for GMMFormer~\cite{gmmformer}). These patterns highlight that fixed clip construction may not adapt to diverse temporal characteristics across videos, motivating the need for a video-specific representation with adaptive temporal granularity. We further provide a direct quantitative analysis of segmentation quality and semantic dilution in Appendix~\ref{sec:seg_analysis}.

\noindent\textbf{Sensitivity to Retrieval Score Weighting.}
The final retrieval score in existing PRVR methods~\cite{ms-sl, gmmformer, hlformer, msc-prvr, proto} is computed as $S = \alpha \cdot S_{\text{clip}} + (1-\alpha) \cdot S_{\text{frame}}$, where $S_{\text{clip}}$ and $S_{\text{frame}}$ are obtained by max-pooling over clip and frame representations, and $\alpha$ balances the two branches. We analyze the sensitivity of retrieval performance to $\alpha$. As shown in Fig.~\ref{fig:anlysis} (a), sweeping $\alpha$ with GMMFormer~\cite{gmmformer} shows that the optimal value differs across datasets (\textit{e.g.}, 0.6 for ActivityNet Captions~\cite{activitynet} and 0.8 for Charades-STA~\cite{charades-sta}), suggesting that the relative importance of two branches depends on dataset-specific temporal characteristics. We further identify the optimal $\alpha$ for each test query as the value yielding the best rank for the ground-truth video. Figure~\ref{fig:anlysis} (b) shows substantial variation in these optima across queries, suggesting that a globally fixed weighting may be suboptimal.

\noindent\textbf{Temporal Inconsistency in Branch-Wise Aggregation.}
Existing methods~\cite{ms-sl, gmmformer, hlformer, msc-prvr, proto} independently aggregate clip- and frame-level similarities to compute the final retrieval score. However, this branch-wise aggregation does not ensure that the selected clip and frame representations correspond to the same temporal region. To examine this, we measure the normalized temporal distance between the clip and frame representations selected for each query, where values close to 0 indicate temporal alignment and values close to 1 indicate misalignment. As shown in Fig.~\ref{fig:anlysis} (c), the distance distribution shows a long-tailed pattern, indicating severe misalignment for a non-negligible number of queries. The same-group rate is only 28.8\%, 48.6\%, and 38.5\% on ActivityNet Captions, Charades-STA, and TVR, respectively, demonstrating that a large portion of selected clip-frame pairs originate from different temporal regions. These results suggest that independent max-pooling aggregation may fail to ensure temporal consistency across branches, limiting the reliability of the final retrieval score.

\section{TF-PRVR: Training-Free Partially Relevant Video Retrieval Framework}
\label{sec:method}

\subsection{Overview}
\label{sec:overview}

\begin{figure}
    \centering
    \includegraphics[width=1\linewidth]{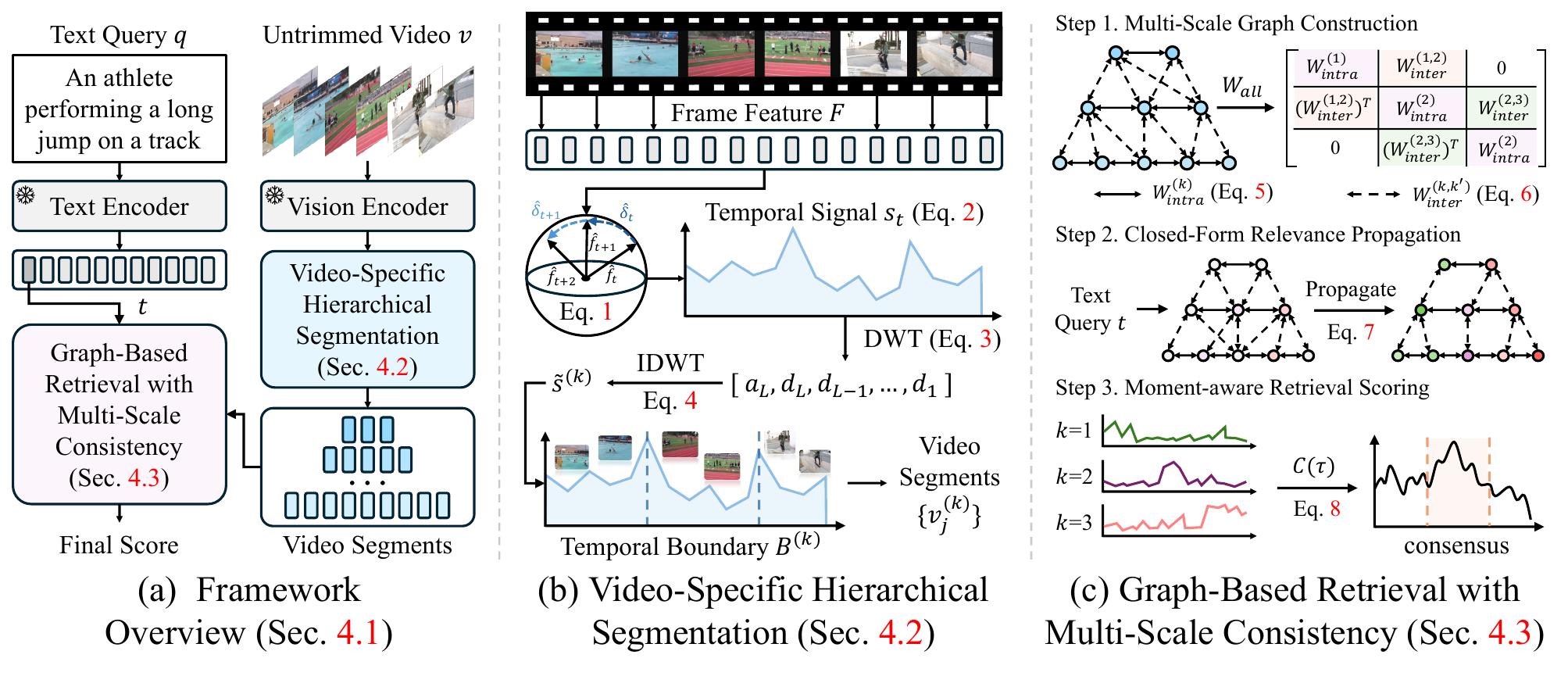}
    \vspace{-1.7em}
\caption{
(a) Overview of TF-PRVR.
(b) Video-specific hierarchical segmentation derives temporal semantic signals from frame features and decomposes them into multi-scale boundaries to construct coherent video segments.
(c) Graph-based retrieval propagates query relevance over a unified multi-scale segment graph and aggregates scale-consistent responses for moment-aware scoring.
}
    \label{fig:main}
    \vspace{-0.5em}
\end{figure}

We propose TF-PRVR, a training-free PRVR framework that avoids source-specific task optimization and addresses rigid video decomposition. Given a text query $q$ and a corpus of $N$ untrimmed videos $\mathcal{V} = \{v_1, v_2, \ldots, v_N\}$, TF-PRVR ranks all videos in $\mathcal{V}$ by their partial relevance to $q$ without any task-specific training. An overview is illustrated in Fig.~\ref{fig:main} (a). TF-PRVR first extracts visual and textual representations using a frozen pre-trained vision-language model~\cite{CLIP, siglip, eva-clip}. For each video, we encode individual frames to obtain frame-level features $F = [f_1, f_2, \ldots, f_{N_f}] \in \mathbb{R}^{N_f \times d}$, where $N_f$ and $d$ denote the number of frames and feature dimension, respectively. For the text query, we combine the [EOS] token embedding with the preceding token embedding to obtain a query representation $t \in \mathbb{R}^{d}$. Instead of constructing a fixed number of clip representations, TF-PRVR adaptively decomposes each video according to its semantic structure. It derives temporal semantic signals from frame features and detects video-specific boundaries across multiple temporal scales, producing hierarchical video segments. TF-PRVR then builds a unified multi-scale graph over these segments, where edges capture temporal and semantic relationships within and across scales. Query-segment similarities are used as initial relevance scores, refined through graph-based propagation, and finally aggregated by moment-aware scoring to produce the video-query score.

\subsection{Video-Specific Hierarchical Segmentation}
\label{sec:seg}
\noindent\textbf{Temporal Signal.}
Given frame features $F$, we first construct a temporal signal that captures semantic transitions across frames. Since relevant moments exhibit temporally coherent visual semantics, identifying changes in the semantic trajectory provides useful cues for locating potential event boundaries. To this end, we measure how the direction of semantic change varies over time. Specifically, we compute the frame-wise velocity vector between consecutive $\ell_2$-normalized frame features:
\begin{equation}
    \delta_t = \hat{f}_{t+1} - \hat{f}_t, \quad t = 1, \ldots, N_f - 1.
\end{equation}
We then define the temporal signal as the cosine dissimilarity between consecutive velocity vectors:
\begin{equation}
    s_t = 1 - \langle \hat{\delta}_t, \hat{\delta}_{t+1} \rangle, \quad t = 1, \ldots, N_f - 2,
\end{equation}
where $\hat{\delta}_t$ denotes the $\ell_2$-normalized velocity at time $t$. A large value of $s_t$ indicates a rapid change in the direction of semantic evolution, suggesting a potential boundary between distinct events.

\noindent\textbf{Multi-Scale Decomposition.}
Since events in untrimmed videos can be defined at varying temporal granularities, a fixed two-level hierarchy may fail to capture video-specific semantic structures. To identify temporal boundaries at multiple granularities in a video-specific manner, we apply the Discrete Wavelet Transform (DWT) to decompose $s$ into frequency components:
\begin{equation}
    [a_L,\ d_L,\ d_{L-1},\ \ldots,\ d_1] = \text{DWT}(s;\ w),
\end{equation}
where $w$ denotes the wavelet basis (\textit{e.g.}, db4), $a_L$ is the low-frequency approximation coefficient, and $d_\ell$ is the detail coefficient at scale $\ell$. Coarser scales (\textit{e.g.}, $d_L$) capture long-range semantic transitions, while finer scales (\textit{e.g.}, $d_1$) correspond to short-range local changes. The number of decomposition levels $L$ varies across videos, naturally yielding a video-specific multi-scale representation.

\noindent\textbf{Temporal Boundary Detection.}
Since the approximation coefficient $a_L$ captures low-frequency trends rather than transitions, we focus on the detail coefficients $\{d_k\}$, which encode semantic changes at each temporal scale. For each detail level $k$, we reconstruct a scale-specific temporal response by applying inverse DWT (IDWT) while retaining only the corresponding detail coefficient $d_k$:
\begin{equation}
    \tilde{s}^{(k)} = \mathrm{IDWT}([0, \ldots, 0,\ d_k,\ 0, \ldots, 0];\ w),
\end{equation}
where $\tilde{s}^{(k)}$ denotes the reconstructed response at scale $k$. We then detect temporal boundaries from local maxima of $|\tilde{s}^{(k)}|$, obtaining a boundary set $\mathcal{B}^{(k)}$ for each scale. Applying this process across all detail levels yields multi-scale boundary sets for video-specific hierarchical segmentation.

\noindent\textbf{Hierarchical Segment Construction.}
At each scale, we combine the detected boundaries with those from coarser scales and partition the video into contiguous segments. This forms a nested hierarchy, where finer segments lie within coarser ones. We average the normalized frame features within each segment to obtain $\{v_j^{(k)}\}_{j=1}^{M_k}$, where $M_k$ varies across videos and scales. We also include individual frames as the finest scale, preserving local visual details alongside broader temporal context.


\subsection{Graph-Based Retrieval with Multi-Scale Consistency}
\label{sec:graph}
We build a single graph from the hierarchical segment representations $\{v_j^{(k)}\}_{j=1}^{M_k}$. The graph connects related segments within and across scales. We then propagate information through the graph and combine segments covering the same time to compute the video-query score.

\noindent\textbf{Intra-Level Affinity.}
Within each scale, we connect segments based on visual similarity and temporal distance. Let $\hat{v}_i^{(k)}$ be the normalized representation of segment $i$ at scale $k$. We define visual affinity as $\phi(u,v)=[\max\{0,\langle u,v\rangle\}]^{\gamma}$ and compute
\begin{equation}
    W_{\text{intra}}^{(k)}[i,j]
    =
    \phi(\hat{v}_i^{(k)},\hat{v}_j^{(k)})
    +
    \lambda_t \exp\!\left(-\frac{|t_i^{(k)}-t_j^{(k)}|}{\tau_k}\right),
\end{equation}
where $t_i^{(k)}$ is the segment center and $\tau_k=\eta\,\overline{\Delta t}^{(k)}$ is proportional to the mean segment duration. The parameters $\gamma$ and $\lambda_t$ control visual affinity and temporal proximity. We keep the strongest connections using nearest-neighbor selection and symmetrize the resulting affinities.

\noindent\textbf{Inter-Level Affinity.}
Segments at different scales can describe the same event over different durations. We connect segments from adjacent scales when they overlap in time and have similar visual content. Let $I_i^{(k)}=[s_i^{(k)},e_i^{(k)})$ denote a segment's temporal span. For $|k-k'|=1$, we define
\begin{equation}
    W_{\text{inter}}[i^{(k)},j^{(k')}]
    =
    \lambda_c
    \frac{|I_i^{(k)}\cap I_j^{(k')}|}
    {\sqrt{|I_i^{(k)}|\,|I_j^{(k')}|}}
    \phi(\hat{v}_i^{(k)},\hat{v}_j^{(k')}),
\end{equation}
where $|I|$ denotes duration and $\lambda_c$ controls the strength of cross-scale connections. The overlap term accounts for differences in segment length. These connections allow short segments to exchange information with broader temporal context.

\noindent\textbf{Unified Multi-Scale Graph.}
We combine the intra- and inter-level affinities into $W_{\text{all}}$. Its diagonal blocks connect segments within each scale, while adjacent off-diagonal blocks connect neighboring scales in both directions. We include self-connections and normalize each row as $W_{\text{norm}}=D^{-1}W_{\text{all}}$, where $D_{ii}=\sum_j W_{\text{all}}[i,j]$. This makes each node combine information from its neighbors with normalized weights.

\noindent\textbf{Closed-Form Relevance Propagation.}
Let $H$ contain the normalized segment representations and $\hat{t}$ be the normalized query representation. Their similarities give the initial relevance scores $y=H\hat{t}$. We refine these scores through graph propagation~\cite{zhou2003learning, belkin2006manifold, yang2026videodetective}:
\begin{equation}
    r^*=(1-\beta)(I-\beta W_{\text{norm}})^{-1}y=\mathcal{K}y,
\end{equation}
where $\beta\in(0,1)$ controls the contribution of neighboring nodes and $\mathcal{K}=(1-\beta)(I-\beta W_{\text{norm}})^{-1}$. Since $r^*=(\mathcal{K}H)\hat{t}$, we precompute $Z=\mathcal{K}H$ for each video. The rows of $Z$ are the propagated segment representations used below.

\noindent\textbf{Moment-Aware Retrieval Scoring.}
We combine segments covering the same time point to obtain a moment-level score. Over a temporal grid $\mathcal{T}$, let $z^{(k)}(\tau)$ and $\ell^{(k)}(\tau)$ denote the propagated representation and duration of the segment covering $\tau$ at scale $k$. We compute
\begin{equation}
    C(\tau)=\left\langle \hat{t},\,\mathcal{N}_{b}\!\left(\frac{\sum_{k\in\mathcal{K}_{\tau}}\ell^{(k)}(\tau)^{-p}\mathcal{N}_{a}\!\left(z^{(k)}(\tau)\right)}{\sum_{k\in\mathcal{K}_{\tau}}\ell^{(k)}(\tau)^{-p}}\right)\right\rangle,
\end{equation}
where $\mathcal{K}_{\tau}$ contains the scales covering $\tau$. The function $\mathcal{N}_{\nu}(x)=x/\|x\|_2^{\nu}$ controls normalization before and after aggregation through $a$ and $b$. The exponent $p$ controls duration weighting: $p=0$ gives equal weights, while $p>0$ gives more weight to shorter segments. The final video score averages the top-$J$ available temporal scores, with $J=1$ giving maximum pooling.

\section{Experiments}
\label{sec:experiments}

\subsection{Experimental Setup}

\noindent\textbf{Datasets and Evaluation Metrics.}
We evaluate TF-PRVR on three standard PRVR benchmarks: ActivityNet Captions~\cite{activitynet}, Charades-STA~\cite{charades-sta}, and TV show Retrieval (TVR)~\cite{tvr}. ActivityNet Captions contains diverse YouTube activity videos with temporally localized descriptions, Charades-STA provides sentence-level annotations for indoor activity segments, and TVR includes videos from six TV shows paired with moment-level natural language queries. Following prior PRVR works~\cite{ms-sl, gmmformer, msc-prvr, hlformer, proto}, we use the standard splits for all datasets. We report rank-based retrieval metrics, $\mathrm{R}@K$ with $K \in \{1,5,10,100\}$, and SumR, the sum $\mathrm{R}@1$, $\mathrm{R}@5$, $\mathrm{R}@10$, and $\mathrm{R}@100$.

\noindent\textbf{Implementation Details.}
TF-PRVR uses frozen vision-language backbones and requires no task-specific training. We apply the same framework across all datasets, with dataset-specific hyperparameters for temporal segmentation, graph propagation, and retrieval scoring. Multi-scale boundaries are obtained through wavelet decomposition, and graph-propagated segment representations are precomputed offline. At inference, each query is compared with the precomputed moment representations to obtain video-level scores. Experiments are conducted on an NVIDIA RTX 6000 Ada GPU with an Intel Xeon Scalable Processor 6526Y CPU.

\subsection{Comparison with the State-of-the-Art}

\begin{table}[t]
\centering
\caption{
Comparison on ActivityNet Captions~\cite{activitynet} under cross-dataset transfer and training-free settings.
Source training data indicates the PRVR training dataset(s).
TF-PRVR$^{\dagger}$ and TF-PRVR$^{\ddagger}$ use CLIP-L/14 and EVA-CLIP-L/14~\cite{eva-clip}, respectively, without source-domain PRVR training.
}
\vspace{0.4em}
\label{tab:cross_domain_act}
\small
\setlength{\tabcolsep}{11pt}

\resizebox{\linewidth}{!}{
\begin{tabular}{llccccc}
\toprule
Source Training Data & Method
& R@1 & R@5 & R@10 & R@100 & SumR \\
\midrule

\rowcolor{gray!12}
\multicolumn{7}{l}{\textit{Training-based methods with CLIP-L~\cite{CLIP}}} \\

\multirow{4}{*}{Charades-STA~\cite{charades-sta}}
& MS-SL~\cite{ms-sl}
& 0.4 & 2.2 & 3.9 & 17.3 & 23.8 \\
& GMMFormer~\cite{gmmformer}
& 0.4 & 2.6 & 4.3 & 18.8 & 26.1 \\
& GMMFormerv2~\cite{gmmformerv2}
& 0.6 & 2.8 & 4.6 & 19.6 & 27.6 \\
& MSC-PRVR~\cite{msc-prvr}
& 0.7 & 2.8 & 4.5 & 20.1 & 28.1 \\

\midrule

\multirow{4}{*}{TVR~\cite{tvr}}
& MS-SL~\cite{ms-sl}
& 6.9 & 19.8 & 26.5 & 59.1 & 112.3 \\
& GMMFormer~\cite{gmmformer}
& 7.2 & 20.6 & 27.1 & 61.2 & 116.1 \\
& GMMFormerv2~\cite{gmmformerv2}
& 8.1 & 21.9 & 30.8 & 62.9 & 123.7 \\
& MSC-PRVR~\cite{msc-prvr}
& 8.0 & 21.1 & 30.1 & 64.6 & 123.8 \\

\midrule

\multirow{4}{*}{Charades-STA~\cite{charades-sta} + TVR~\cite{tvr}}
& MS-SL~\cite{ms-sl}
& 5.4 & 15.8 & 21.2 & 58.9 & 101.3 \\
& GMMFormer~\cite{gmmformer}
& 5.9 & 16.8 & 22.4 & 60.0 & 105.1 \\
& GMMFormerv2~\cite{gmmformerv2}
& 5.6 & 16.2 & 20.9 & 60.8 & 103.5 \\
& MSC-PRVR~\cite{msc-prvr}
& 6.1 & 17.1 & 24.8 & 58.3 & 106.3 \\

\midrule
\rowcolor{gray!12}
\multicolumn{7}{l}{\textit{Training-free methods}} \\

\multirow{2}{*}{None}
& \textbf{TF-PRVR$^\dagger$ (Ours)}
& 17.5 & 35.1 & 55.7 & 85.6 & 193.9 \\
& \textbf{TF-PRVR$^\ddagger$ (Ours)}
& 18.8 & 37.3 & 56.9 & 87.8 & 200.8 \\

\bottomrule
\end{tabular}
}
\end{table}

\noindent\textbf{Cross-Dataset Transfer Comparison.}
We examine the transfer degradation of existing training-based PRVR methods by training them on source datasets and evaluating them on ActivityNet Captions~\cite{activitynet}. As shown in Table~\ref{tab:cross_domain_act}, training-based methods suffer substantial degradation when transferred to the unseen target dataset. For example, MSC-PRVR~\cite{msc-prvr} obtains SumR scores of 28.1, 123.8, and 106.3 when trained on Charades-STA~\cite{charades-sta}, TVR~\cite{tvr}, and their combination, respectively, compared with 202.1 under in-domain ActivityNet training (Table~\ref{tab:in-domain}). These results suggest that task-specific PRVR optimization can lead to substantial source dependence. In contrast, TF-PRVR$^{\ddagger}$ achieves a SumR of 200.8 on ActivityNet Captions without any source-domain PRVR training, exceeding all source-trained baselines evaluated on the same target dataset. This result demonstrates effective training-free retrieval without relying on source-specific optimization. Additional comparisons on Charades-STA and TVR are provided in Tables~\ref{tab:cross_domain_cha} and~\ref{tab:cross_domain_tvr}, respectively.

\noindent\textbf{In-Domain and Training-Free Comparison.}
We compare TF-PRVR with existing state-of-the-art PRVR methods on ActivityNet Captions, Charades-STA, and TVR. Training-based methods are evaluated on the same dataset used for training, whereas TF-PRVR requires no PRVR-specific training. As shown in Table~\ref{tab:in-domain}, TF-PRVR achieves competitive performance across all three benchmarks. On ActivityNet Captions, TF-PRVR obtains an R@1 of 17.5, comparable to MSC-PRVR~\cite{msc-prvr} at 17.7, indicating that video-specific hierarchical segmentation and graph-based relevance propagation can effectively exploit frozen vision-language representations without target-specific optimization. TF-PRVR also improves over the CLIP-L zero-shot baseline, increasing SumR from 172.9 to 193.9 on ActivityNet Captions, from 73.1 to 80.8 on Charades-STA, and from 167.2 to 178.3 on TVR. Notably, several training-based methods with CNN-based visual backbones, such as ResNet152+I3D, underperform CLIP-L zero-shot on ActivityNet Captions and TVR, suggesting that strong vision-language alignment can provide substantial benefits even without task-specific training. Although training-based methods achieve higher overall in-domain performance, TF-PRVR remains competitive without PRVR-specific optimization, highlighting the effectiveness of our training-free design.

\begin{table}[t]
\centering
\vspace{-1.0em}
\caption{
Comparison with state-of-the-art PRVR methods on ActivityNet Captions~\cite{activitynet},
Charades-STA~\cite{charades-sta}, and TVR~\cite{tvr}.
Training-based methods are evaluated in-domain, while training-free methods require no PRVR-specific training.
Methods are grouped by training setting and backbone.
}
\vspace{0.3em}
\label{tab:in-domain}
\scriptsize
\setlength{\tabcolsep}{2.5pt}
\resizebox{\linewidth}{!}{
\begin{tabular}{lccccc ccccc ccccc}
\toprule
\multirow{2}{*}{Method}
& \multicolumn{5}{c}{ActivityNet Captions~\cite{activitynet}}
& \multicolumn{5}{c}{Charades-STA~\cite{charades-sta}}
& \multicolumn{5}{c}{TVR~\cite{tvr}} \\
\cmidrule(lr){2-6} \cmidrule(lr){7-11} \cmidrule(lr){12-16}
& R@1 & R@5 & R@10 & R@100 & SumR
& R@1 & R@5 & R@10 & R@100 & SumR
& R@1 & R@5 & R@10 & R@100 & SumR \\
\midrule
\midrule
\rowcolor{gray!12}
\multicolumn{16}{l}{\textit{Training-based methods with ResNet152~\cite{resnet} + I3D~\cite{i3d} + RoBERTa~\cite{roberta}}} \\
MS-SL~\cite{ms-sl}
& 7.1 & 22.5 & 34.7 & 75.8 & 140.1
& 1.8 & 7.1 & 11.8 & 47.7 & 68.4
& 13.5 & 32.1 & 43.4 & 83.4 & 172.4 \\
PEAN~\cite{jiang2023progressive}
& 7.4 & 23.0 & 35.5 & 75.9 & 141.8
& 2.7 & 8.1 & 13.5 & 50.3 & 74.7
& 13.5 & 32.8 & 44.1 & 83.9 & 174.2 \\
LH~\cite{fang2024linguistic}
& 7.4 & 23.5 & 35.8 & 75.8 & 142.4
& 2.1 & 7.5 & 12.9 & 50.1 & 72.7
& 13.2 & 33.2 & 44.4 & 85.5 & 176.3 \\
BGM-Net~\cite{yin2024exploiting}
& 7.2 & 23.8 & 36.0 & 76.9 & 143.9
& 1.9 & 7.4 & 12.2 & 50.1 & 71.6
& 14.1 & 34.7 & 45.9 & 85.2 & 179.9 \\
GMMFormer~\cite{gmmformer}
& 8.3 & 24.9 & 36.7 & 76.1 & 146.0
& 2.1 & 7.8 & 12.5 & 50.6 & 72.9
& 13.9 & 33.3 & 44.5 & 84.9 & 176.6 \\
DL-DKD~\cite{dl-dkd}
& 8.0 & 25.0 & 37.5 & 77.1 & 147.6
& - & - & - & - & -
& 14.4 & 34.9 & 45.8 & 84.9 & 179.9 \\
GMMFormerv2~\cite{gmmformerv2}
& 8.9 & 27.1 & 40.2 & 78.7 & 154.9
& 2.5 & 8.6 & 13.9 & 53.2 & 78.2
& 16.2 & 37.6 & 48.8 & 86.5 & 189.1 \\
ProtoPRVR~\cite{proto}
& 7.9 & 24.9 & 37.2 & 77.3 & 147.4
& - & - & - & - & -
& 15.4 & 35.9 & 47.5 & 86.4 & 185.2 \\
HLFormer~\cite{hlformer}
& 8.7 & 27.1 & 40.1 & 79.0 & 154.9
& 2.6 & 8.5 & 13.7 & 54.0 & 78.7
& 15.7 & 37.1 & 48.5 & 86.4 & 187.7 \\

\midrule
\rowcolor{gray!12}
\multicolumn{16}{l}{\textit{Training-based methods with CLIP-L~\cite{CLIP}}} \\
MS-SL~\cite{ms-sl}
& 14.7 & 37.1 & 50.4 & 84.6 & 186.7
& 3.4 & 11.5 & 18.7 & 62.5 & 96.0
& 31.9 & 57.6 & 67.7 & 93.8 & 251.0 \\
GMMFormer~\cite{gmmformer}
& 15.2 & 37.7 & 50.5 & 83.7 & 187.1
& 2.7 & 10.5 & 16.7 & 59.4 & 89.3
& 29.8 & 54.2 & 64.6 & 92.5 & 241.1 \\
AMDNet~\cite{song2025towards}
& 14.0 & 36.3 & 49.9 & 84.2 & 184.5
& 2.1 & 7.8 & 13.9 & 57.2 & 81.1
& 27.7 & 52.3 & 63.3 & 92.3 & 235.6 \\
BGM-Net~\cite{yin2024exploiting}
& 15.6 & 37.9 & 51.3 & 85.4 & 190.3
& 3.0 & 11.8 & 18.2 & 63.7 & 96.7
& 31.1 & 56.3 & 66.5 & 93.8 & 247.7 \\
GMMFormerv2~\cite{gmmformerv2}
& 17.1 & 40.6 & 53.7 & 85.5 & 196.9
& 3.1 & 11.6 & 18.2 & 61.4 & 94.2
& 34.0 & 59.7 & 69.8 & 94.6 & 258.1 \\
ProtoPRVR~\cite{proto}
& 16.0 & 38.8 & 52.4 & 85.1 & 192.3
& - & - & - & - & -
& 34.7 & 60.0 & 70.1 & 94.4 & 259.2 \\
ARL~\cite{arl}
& 15.3 & 38.4 & 51.5 & 85.2 & 190.4
& - & - & - & - & -
& 34.6 & 60.4 & 70.7 & 94.4 & 260.1 \\
MSC-PRVR~\cite{msc-prvr}
& 17.7 & 42.0 & 55.6 & 86.8 & 202.1
& 3.2 & 12.6 & 20.1 & 63.8 & 99.7
& 35.1 & 61.6 & 71.5 & 94.9 & 263.1 \\

\midrule
\rowcolor{gray!12}
\multicolumn{16}{l}{\textit{Training-free methods with CLIP-L~\cite{CLIP}}} \\
CLIP-L Zero-shot~\cite{CLIP}
& 15.1  & 33.9 & 45.1 & 78.9 & 172.9
& 2.0 & 8.1 & 13.6 & 49.4 & 73.1
& 16.2  & 33.5  & 41.8  & 75.7 & 167.2 \\
\textbf{TF-PRVR (Ours)}
& 17.5 & 35.1 & 55.7 & 85.6 & 193.9 
& 2.6 & 9.1 & 15.5 & 53.6 & 80.8
& 18.0 & 35.8 & 45.4 & 79.1 & 178.3 \\
\bottomrule
\end{tabular}
}
\end{table}

\subsection{Ablation Study}

\begin{table}[t]
\centering
\vspace{-0.5em}
\caption{Ablation studies on ActivityNet Captions~\cite{activitynet}. HS, GP, and MAS denote hierarchical segmentation, graph-based relevance propagation, and moment-aware scoring, respectively.}
\vspace{-0.2em}
\label{tab:ablation_studies}
\scriptsize
\setlength{\tabcolsep}{3pt}

\begin{subtable}[t]{0.30\textwidth}
\centering
\caption{Backbone ablation.}
\vspace{-0.5em}
\label{tab:backbone_ablation}
{
\renewcommand{\arraystretch}{1.0}
\resizebox{\linewidth}{!}{
\begin{tabular}{ll|c}
\toprule
Backbone & Method & SumR \\
\midrule
CLIP-L/14~\cite{CLIP} & Zero-shot & 172.9 \\
CLIP-L/14~\cite{CLIP} & TF-PRVR & 193.9 \\
\midrule
SigLIP-L/14~\cite{siglip} & Zero-shot & 178.7 \\
SigLIP-L/14~\cite{siglip} & TF-PRVR & 198.5 \\
\midrule
EVA-CLIP-L/14~\cite{eva-clip} & Zero-shot & 180.4  \\
EVA-CLIP-L/14~\cite{eva-clip} & TF-PRVR & 200.8 \\
\bottomrule
\end{tabular}
}
}
\end{subtable}
\hfill
\begin{subtable}[t]{0.36\textwidth}
\centering
\caption{Module components ablation.}
\vspace{-0.5em}
\label{tab:module_ablation}
{
\renewcommand{\arraystretch}{1.29}
\resizebox{\linewidth}{!}{
\begin{tabular}{lccc|c}
\toprule
Method & HS & GP & MAS & SumR \\
\midrule
EVA-CLIP-L/14~\cite{eva-clip} & - & - & - & 180.4 \\
+ Fixed Seg. & - & - & - & 186.6 \\
+ HS & \checkmark & - & - & 194.7 \\
+ HS + GP & \checkmark & \checkmark & - & 197.8 \\
TF-PRVR & \checkmark & \checkmark & \checkmark & 200.8 \\
\bottomrule
\end{tabular}
}
}
\end{subtable}
\hfill
\begin{subtable}[t]{0.31\textwidth}
\centering
\caption{Temporal signal ablation.}
\vspace{-0.5em}
\label{tab:signal_ablation}
{
\renewcommand{\arraystretch}{1.43}
\resizebox{\linewidth}{!}{
\begin{tabular}{lc}
\toprule
Temporal Signal & SumR \\
\midrule
Uniform fixed intervals & 191.8 \\
Adjacent feature dissimilarity & 196.5 \\
Self-similarity matrix & 197.6 \\
Velocity-direction change & 200.8 \\
\bottomrule
\end{tabular}
}
}
\end{subtable}
\end{table}
\vspace{-0.2em}

\noindent\textbf{Backbone Ablation.}
We examine whether TF-PRVR consistently improves retrieval performance across different frozen vision-language backbones. As shown in Table~\ref{tab:ablation_studies} (a), TF-PRVR substantially improves over the corresponding zero-shot baseline for all backbones. Specifically, TF-PRVR increases SumR from 172.9 to 193.9 with CLIP-L/14~\cite{CLIP}, from 178.7 to 198.5 with SigLIP-L/14~\cite{siglip}, and from 180.4 to 200.8 with EVA-CLIP-L/14~\cite{eva-clip}. These consistent gains indicate that TF-PRVR is not tied to a specific backbone. Since our modules operate directly on frozen frame and text features without any task-specific optimization, they can be applied in a plug-and-play manner to different pre-trained vision-language models, demonstrating the flexibility of our training-free framework.

\noindent\textbf{Module Component Ablation.}
We analyze the contribution of each module in TF-PRVR. As shown in Table~\ref{tab:ablation_studies} (b), we start from the zero-shot EVA-CLIP-L/14~\cite{eva-clip} baseline, which scores videos by max-pooling frame-query similarities. \textit{Fixed Seg.} follows the conventional two-branch setting: it uniformly constructs fixed clip-level segments in addition to frame-level representations, independently applies max pooling in both branches, and combines their scores. This improves SumR from 180.4 to 186.6, but remains limited by content-agnostic decomposition. Replacing it with hierarchical segmentation (HS) increases SumR to 194.7, showing the benefit of video-specific multi-scale semantic boundaries. Adding graph-based relevance propagation (GP) further refines query-segment relevance over the multi-scale graph, improving SumR to 197.8. Finally, moment-aware scoring (MAS) aggregates temporally aligned relevance across scales, achieving the best SumR of 200.8. These results show that HS reduces semantic dilution, while GP and MAS enhance multi-scale temporal consistency.

\noindent\textbf{Temporal Signal Ablation.}
We further analyze the effect of different temporal signals for constructing video-specific boundaries. In Table~\ref{tab:ablation_studies} (c), all other modules are fixed, and only the temporal signal construction strategy is varied. \textit{Uniform fixed intervals} divide videos without visual content, and thus cannot adapt to video-specific semantic transitions. \textit{Adjacent feature dissimilarity} measures cosine dissimilarity between consecutive frame features, but can be sensitive to local appearance changes. \textit{Self-similarity matrix} derives boundary responses from changes in local similarity patterns within a temporal window, but requires a predefined window size and may produce overly smooth responses. In contrast, our \textit{velocity-direction change} signal captures changes in the direction of semantic evolution, making it more effective for detecting transitions between distinct events. As shown in Table~\ref{tab:ablation_studies} (c), it achieves the best SumR of 200.8, outperforming the other variants by 3.2--9.0 points. Therefore, we adopt velocity-direction change as the default temporal signal in TF-PRVR. Additional ablation studies and qualitative results are provided in the supplementary material.

\begin{figure}
    \centering
    \includegraphics[width=\linewidth]{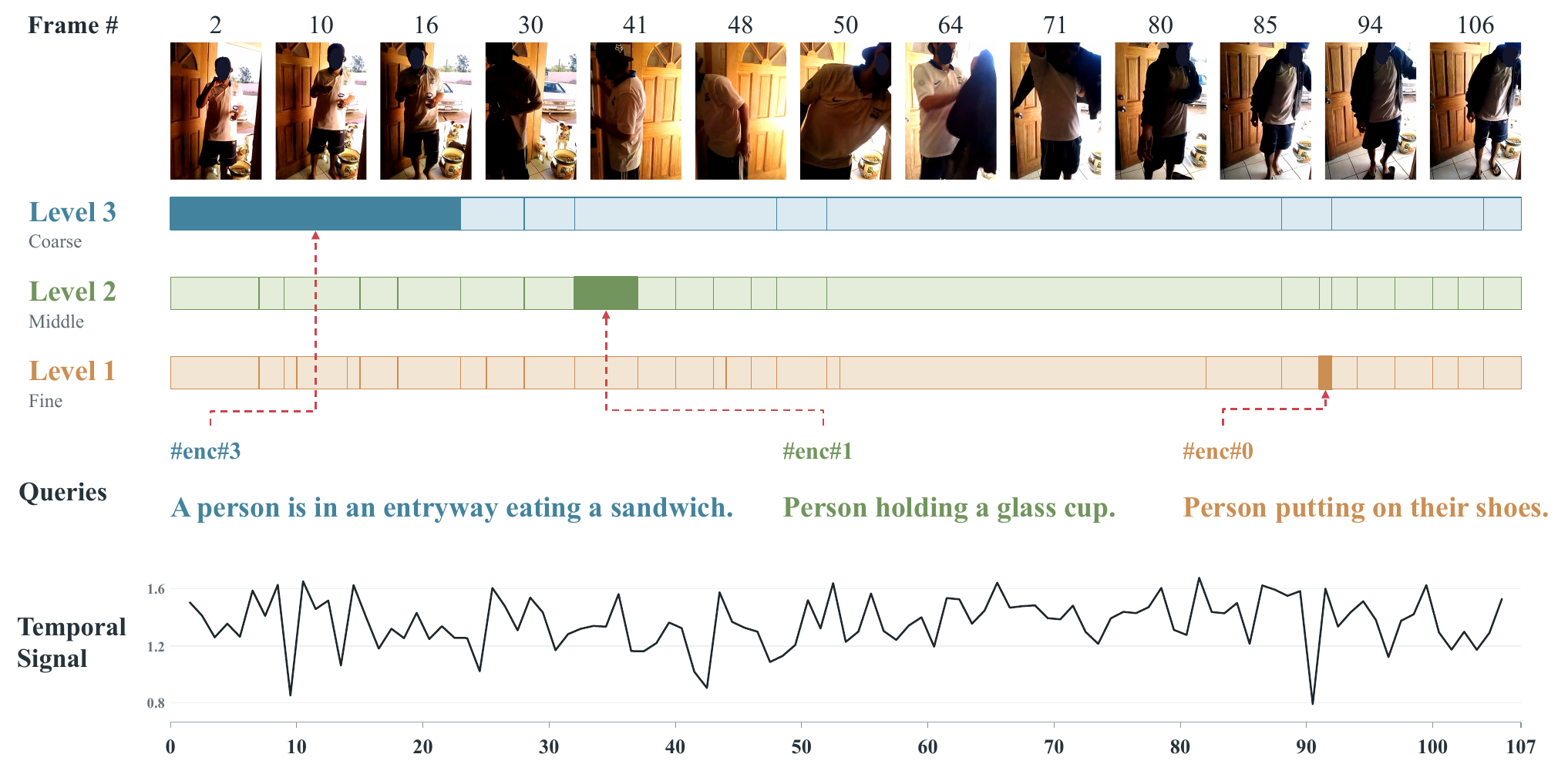}
    \vspace{-1.5em}
\caption{Qualitative visualization of video-specific hierarchical segmentation on Charades-STA~\cite{charades-sta}. The rows show segments at different DWT levels, while the temporal signal is shown below. Red dashed lines connect example queries to their best-matched segments.}
    \label{fig:vis}
    \vspace{-0.4em}
\end{figure}

\subsection{Qualitative Result}
Figure~\ref{fig:vis} visualizes the video-specific hierarchical segmentation produced by our DWT-based decomposition on a representative Charades-STA video (\texttt{9JZO2}, 107 frames). Each level captures temporal structure at a different granularity: finer levels (Level 1) produce shorter segments that isolate local events, while coarser levels (Level 3) capture broader temporal context. The red dashed lines show that the example text queries align with different segments across levels, illustrating how the hierarchy can capture query-relevant content at appropriate temporal scales. In contrast, fixed clip construction would aggregate approximately 3 frames per clip for this video, fragmenting relevant moments across multiple clips and potentially mixing them with unrelated content. Our video-specific segmentation instead adapts the temporal granularity to the underlying video structure, producing more semantically coherent segments and reducing the risk of semantic dilution.

\section{Limitations and Future Work}
While TF-PRVR achieves consistent training-free retrieval across diverse datasets, its performance depends on the zero-shot capability of the frozen backbone and may be limited in specialized domains. It also requires offline graph computation and storage of moment-level representations, introducing preprocessing and indexing costs. These representations are computed once and reused across queries. Future work may address these limitations through stronger frozen backbones and more efficient graph computation and moment indexing.

\section{Conclusion}
We proposed TF-PRVR, a training-free framework for partially relevant video retrieval that addresses source dependence and rigid video decomposition. Using frozen vision-language features, TF-PRVR constructs video-specific temporal hierarchies and combines multi-scale graph propagation with moment-aware scoring to integrate temporally aligned evidence. Experiments demonstrate its effectiveness across datasets with diverse visual and temporal characteristics without task-specific model training, supporting a practical direction to training-free PRVR.

\clearpage

\section*{Acknowledgements}
This work was supported by the National Research Foundation of Korea (NRF) grant funded by the Korea government (MSIT) (RS-2024-00355008), the Institute of Information \& communications Technology Planning \& Evaluation (IITP) grant funded by the Korea government (MSIT) (RS-2025-25422680), and the Startup Technology Development Program (RS-2026-25513099) funded by the Ministry of SMEs and Startups (MSS, Korea).

{
    \small
    \bibliographystyle{abbrv}
    \bibliography{reference}
}
\newpage
\clearpage

\FloatBarrier

\begin{center}
    {\Large \textbf{Appendix}}
\end{center}

\vspace{1em}

\renewcommand{\thesection}{A}   
\renewcommand{\thetable}{A\arabic{table}}   
\renewcommand{\thefigure}{A\arabic{figure}}
\setcounter{section}{0}
\setcounter{table}{0}
\setcounter{figure}{0}

\begin{table}[h]
\centering
\caption{
Comparison on additional video datasets.
Source training data denotes the PRVR dataset(s) used to train each method.
TF-PRVR and CLIP zero-shot require no PRVR-specific training.
}
\vspace{0.4em}
\label{tab:additional_datasets}
\footnotesize
\setlength{\tabcolsep}{4pt}

\resizebox{\linewidth}{!}{
\begin{tabular}{lllccccc}
\toprule
Dataset & Method & Source Training Data
& R@1 & R@5 & R@10 & R@100 & SumR \\
\midrule

\multirow{4}{*}{QVHighlights~\cite{qvhighlight}}
& GMMFormerv2~\cite{gmmformerv2}
& Act+Cha+TVR
& 18.6 & 26.5 & 29.8 & 50.6 & 125.5 \\
& MSC-PRVR~\cite{msc-prvr}
& Act+Cha+TVR
& 20.1 & 29.2 & 33.0 & 49.8 & 132.1 \\
& CLIP Zero-shot~\cite{CLIP}
& None
& 21.0 & 41.4 & 52.3 & 82.9 & 197.6 \\
& \textbf{TF-PRVR (Ours)}
& None
& 23.9 & 47.5 & 57.6 & 89.5 & 218.5 \\

\midrule

\multirow{4}{*}{EPIC-Kitchens-100~\cite{epic100}}
& GMMFormerv2~\cite{gmmformerv2}
& Act+Cha+TVR
& 2.5 & 8.1 & 12.4 & 21.1 & 44.1 \\
& MSC-PRVR~\cite{msc-prvr}
& Act+Cha+TVR
& 2.8 & 8.5 & 13.1 & 22.9 & 47.3 \\
& CLIP Zero-shot~\cite{CLIP}
& None
& 4.0 & 14.4 & 23.6 & 36.2 & 78.2 \\
& \textbf{TF-PRVR (Ours)}
& None
& 5.2 & 17.8 & 28.0 & 42.3 & 93.3 \\

\midrule

\multirow{4}{*}{TACoS~\cite{tacos}}
& GMMFormerv2~\cite{gmmformerv2}
& Act+Cha+TVR
& 16.8 & 27.5 & 42.1 & 65.0 & 151.4 \\
& MSC-PRVR~\cite{msc-prvr}
& Act+Cha+TVR
& 18.6 & 30.1 & 42.8 & 65.2 & 156.7 \\
& CLIP Zero-shot~\cite{CLIP}
& None
& 30.6 & 41.8 & 52.0 & 75.5 & 199.9 \\
& \textbf{TF-PRVR (Ours)}
& None
& 34.4 & 46.5 & 57.2 & 85.2 & 223.3 \\

\midrule

\multirow{4}{*}{YouCook2~\cite{youcook2}}
& GMMFormerv2~\cite{gmmformerv2}
& Act+Cha+TVR
& 8.4 & 24.1 & 35.6 & 55.0 & 123.1 \\
& MSC-PRVR~\cite{msc-prvr}
& Act+Cha+TVR
& 8.9 & 25.3 & 36.8 & 57.2 & 128.2 \\
& CLIP Zero-shot~\cite{CLIP}
& None
& 12.5 & 34.5 & 48.0 & 75.4 & 170.4 \\
& \textbf{TF-PRVR (Ours)}
& None
& 14.2 & 38.0 & 53.8 & 79.6 & 185.6 \\

\bottomrule
\end{tabular}
}
\end{table}
\renewcommand{\thesection}{A}
\section{More Results}

\subsection{Evaluation on Additional Diverse Video Datasets}
\label{sec:additional_datasets}

To further evaluate TF-PRVR across diverse visual and temporal characteristics, we conduct additional experiments on QVHighlights~\cite{qvhighlight}, EPIC-Kitchens-100~\cite{epic100}, TACoS~\cite{tacos}, and YouCook2~\cite{youcook2}. These datasets cover diverse consumer videos, egocentric kitchen activities, fine-grained cooking actions, and procedural instructional videos, respectively. As shown in Table~\ref{tab:additional_datasets}, TF-PRVR consistently improves over the zero-shot baseline across all four datasets. Specifically, it improves SumR by 20.9, 15.1, 23.4, and 15.2 on QVHighlights, EPIC-Kitchens-100, TACoS, and YouCook2, respectively. TF-PRVR also outperforms the source-trained PRVR baselines on all four target datasets. These results provide broader evidence that TF-PRVR remains effective across datasets with diverse visual content and temporal characteristics without PRVR-specific training.

\begin{table}[t]
\centering
\caption{
Comparison on Charades-STA~\cite{charades-sta} under in-domain, cross-dataset transfer, and training-free settings.
Source training data indicates the PRVR training dataset(s).
TF-PRVR$^{\dagger}$ and TF-PRVR$^{\ddagger}$ use CLIP-L/14 and EVA-CLIP-L/14~\cite{eva-clip}, respectively, without source-domain PRVR training.
}
\vspace{0.4em}
\label{tab:cross_domain_cha}
\small
\setlength{\tabcolsep}{8pt}

\resizebox{\linewidth}{!}{
\begin{tabular}{llccccc}
\toprule
Source Training Data & Method
& R@1 & R@5 & R@10 & R@100 & SumR \\
\midrule

\rowcolor{gray!12}
\multicolumn{7}{l}{\textit{Training-based methods with CLIP-L~\cite{CLIP}}} \\

\multirow{4}{*}{Charades-STA~\cite{charades-sta}}
& MS-SL~\cite{ms-sl}
& 3.4 & 11.5 & 18.7 & 62.5 & 96.0 \\
& GMMFormer~\cite{gmmformer}
& 2.7 & 10.5 & 16.7 & 59.4 & 89.3 \\
& GMMFormerv2~\cite{gmmformerv2}
& 3.1 & 11.6 & 18.2 & 61.4 & 94.2 \\
& MSC-PRVR~\cite{msc-prvr}
& 3.2 & 12.6 & 20.1 & 63.8 & 99.7 \\

\midrule

\multirow{4}{*}{ActivityNet Captions~\cite{activitynet}}
& MS-SL~\cite{ms-sl}
& 0.8 & 3.4 & 6.2 & 31.7 & 42.1 \\
& GMMFormer~\cite{gmmformer}
& 0.9 & 3.7 & 6.7 & 33.2 & 44.5 \\
& GMMFormerv2~\cite{gmmformerv2}
& 1.0 & 3.9 & 6.9 & 34.0 & 45.8 \\
& MSC-PRVR~\cite{msc-prvr}
& 1.1 & 4.1 & 7.3 & 35.2 & 47.7 \\

\midrule

\multirow{4}{*}{TVR~\cite{tvr}}
& MS-SL~\cite{ms-sl}
& 1.3 & 5.9 & 10.4 & 22.2 & 39.8 \\
& GMMFormer~\cite{gmmformer}
& 1.5 & 6.4 & 11.0 & 23.7 & 42.6 \\
& GMMFormerv2~\cite{gmmformerv2}
& 1.6 & 6.8 & 11.6 & 24.2 & 44.2 \\
& MSC-PRVR~\cite{msc-prvr}
& 1.9 & 7.5 & 12.8 & 25.2 & 47.4 \\

\midrule

\multirow{4}{*}{ActivityNet Captions~\cite{activitynet} + TVR~\cite{tvr}}
& MS-SL~\cite{ms-sl}
& 1.7 & 6.3 & 10.2 & 44.2 & 62.4 \\
& GMMFormer~\cite{gmmformer}
& 1.9 & 6.7 & 10.9 & 45.6 & 65.1 \\
& GMMFormerv2~\cite{gmmformerv2}
& 2.0 & 6.9 & 11.3 & 46.5 & 66.7 \\
& MSC-PRVR~\cite{msc-prvr}
& 2.2 & 7.2 & 11.8 & 47.3 & 68.5 \\

\midrule
\rowcolor{gray!12}
\multicolumn{7}{l}{\textit{Training-free methods}} \\

\multirow{2}{*}{None}
& \textbf{TF-PRVR$^\dagger$ (Ours)}
& 2.6 & 9.1 & 15.5 & 53.6 & 80.8 \\
& \textbf{TF-PRVR$^\ddagger$ (Ours)}
& 2.9 & 10.6 & 16.8 & 55.1 & 85.4 \\

\bottomrule
\end{tabular}
}
\end{table}
\begin{table}[t]
\centering
\caption{
Comparison on TVR~\cite{tvr} under in-domain, cross-dataset transfer, and training-free settings.
Source training data indicates the PRVR training dataset(s).
TF-PRVR$^{\dagger}$ and TF-PRVR$^{\ddagger}$ use CLIP-L/14 and EVA-CLIP-L/14~\cite{eva-clip}, respectively, without source-domain PRVR training.
}
\label{tab:cross_domain_tvr}
\vspace{0.4em}
\small
\setlength{\tabcolsep}{6pt}

\resizebox{\linewidth}{!}{
\begin{tabular}{llccccc}
\toprule
Source Training Data & Method
& R@1 & R@5 & R@10 & R@100 & SumR \\
\midrule

\rowcolor{gray!12}
\multicolumn{7}{l}{\textit{Training-based methods with CLIP-L~\cite{CLIP}}} \\

\multirow{4}{*}{TVR~\cite{tvr}}
& MS-SL~\cite{ms-sl}
& 31.9 & 57.6 & 67.7 & 93.8 & 251.0 \\
& GMMFormer~\cite{gmmformer}
& 29.8 & 54.2 & 64.6 & 92.5 & 241.1 \\
& GMMFormerv2~\cite{gmmformerv2}
& 34.0 & 59.7 & 69.8 & 94.6 & 258.1 \\
& MSC-PRVR~\cite{msc-prvr}
& 35.1 & 61.6 & 71.5 & 94.9 & 263.1 \\

\midrule

\multirow{4}{*}{ActivityNet Captions~\cite{activitynet}}
& MS-SL~\cite{ms-sl}
& 3.8 & 9.8 & 14.2 & 40.7 & 68.5 \\
& GMMFormer~\cite{gmmformer}
& 4.1 & 10.5 & 15.1 & 42.1 & 71.8 \\
& GMMFormerv2~\cite{gmmformerv2}
& 4.6 & 11.3 & 15.9 & 43.8 & 75.6 \\
& MSC-PRVR~\cite{msc-prvr}
& 4.9 & 11.7 & 16.5 & 44.3 & 77.4 \\

\midrule

\multirow{4}{*}{Charades-STA~\cite{charades-sta}}
& MS-SL~\cite{ms-sl}
& 0.3 & 1.7 & 3.1 & 17.3 & 22.4 \\
& GMMFormer~\cite{gmmformer}
& 0.4 & 1.9 & 3.4 & 18.0 & 23.7 \\
& GMMFormerv2~\cite{gmmformerv2}
& 0.5 & 2.0 & 3.7 & 18.7 & 24.9 \\
& MSC-PRVR~\cite{msc-prvr}
& 0.5 & 2.2 & 3.9 & 19.0 & 25.6 \\

\midrule

\multirow{4}{*}{ActivityNet Captions~\cite{activitynet} + Charades-STA~\cite{charades-sta}}
& MS-SL~\cite{ms-sl}
& 3.6 & 9.4 & 13.6 & 39.5 & 66.1 \\
& GMMFormer~\cite{gmmformer}
& 3.9 & 10.1 & 14.4 & 41.0 & 69.4 \\
& GMMFormerv2~\cite{gmmformerv2}
& 4.2 & 10.8 & 15.1 & 41.8 & 71.9 \\
& MSC-PRVR~\cite{msc-prvr}
& 4.6 & 11.3 & 15.5 & 42.3 & 73.7 \\

\midrule
\rowcolor{gray!12}
\multicolumn{7}{l}{\textit{Training-free methods}} \\

\multirow{2}{*}{None}
& \textbf{TF-PRVR$^\dagger$ (Ours)}
& 18.0 & 35.8 & 45.4 & 79.1 & 178.3 \\
& \textbf{TF-PRVR$^\ddagger$ (Ours)}
& 19.2 & 37.1 & 46.8 & 80.5 & 183.6 \\

\bottomrule
\end{tabular}
}
\end{table}

\subsection{Additional Cross-Dataset Transfer Results}

\noindent\textbf{Charades-STA.} We provide additional comparisons on Charades-STA~\cite{charades-sta} under in-domain, cross-dataset transfer, and training-free settings. As shown in Table~\ref{tab:cross_domain_cha}, training-based methods exhibit substantial performance degradation when transferred to Charades-STA from other source datasets. For example, MSC-PRVR~\cite{msc-prvr} achieves a SumR of 99.7 when trained in-domain on Charades-STA, but decreases to 47.7 and 47.4 when trained on ActivityNet Captions~\cite{activitynet} and TVR, respectively. In contrast, TF-PRVR requires no source-domain PRVR training and achieves a SumR of 85.4 with EVA-CLIP-L/14, exceeding all source-trained baselines evaluated on Charades-STA. These results further illustrate the source dependence introduced by task-specific PRVR optimization.

\noindent\textbf{TVR.} Table~\ref{tab:cross_domain_tvr} shows the results on TVR~\cite{tvr}. MSC-PRVR achieves a SumR of 263.1 when trained in-domain on TVR, but drops to 77.4 and 25.6 when trained on ActivityNet Captions and Charades-STA, respectively. Training on both source datasets yields a SumR of 73.7, which remains substantially below the in-domain result. TF-PRVR does not surpass models trained directly on TVR, but achieves a SumR of 183.6 with EVA-CLIP-L/14 without any source-domain PRVR training, exceeding all source-trained transfer baselines on the same target dataset. Together with the results on ActivityNet Captions, these results demonstrate effective training-free retrieval across multiple target datasets without source-specific PRVR optimization.

\renewcommand{\thesection}{B}
\section{Additional Ablation Study}

\begin{table}[t]
\centering
\caption{
Direct analysis of segmentation quality and semantic dilution.
Higher values are better for Mean Best tIoU, R@0.5, and GT Coverage, while lower values are better for Fragmentation, Irrelevant Inclusion, and feature dispersion $D_{\mathrm{all}}$.
GT moments are used only for post-hoc evaluation.
}
\vspace{0.5em}
\label{tab:seg_quality}
\footnotesize
\setlength{\tabcolsep}{5.0pt}

\resizebox{\linewidth}{!}{
\begin{tabular}{llcccccc}
\toprule
Dataset & Segmentation
& \makecell{Mean Best\\tIoU $\uparrow$}
& R@0.5 $\uparrow$
& \makecell{GT\\Coverage $\uparrow$}
& Fragmentation $\downarrow$
& \makecell{Irrelevant\\Inclusion $\downarrow$}
& $D_{\mathrm{all}} \downarrow$ \\
\midrule

\multirow{2}{*}{ActivityNet Captions~\cite{activitynet}}
& Fixed scheme
& 0.177 & 7.2\% & 0.187 & 12.41 & 0.082 & 0.0864 \\
& \textbf{Ours}
& 0.315 & 24.8\% & 0.354
& 5.86 & 0.041 & 0.0592 \\

\midrule

\multirow{2}{*}{Charades-STA~\cite{charades-sta}}
& Fixed scheme
& 0.139 & 0.1\% & 0.151 & 9.78 & 0.057 & 0.0109 \\
& \textbf{Ours}
& 0.278 & 14.2\% & 0.291
& 4.85 & 0.031 & 0.0076 \\

\midrule

\multirow{2}{*}{TVR~\cite{tvr}}
& Fixed scheme
& 0.348 & 27.0\% & 0.367 & 5.37 & 0.076 & 0.0117 \\
& \textbf{Ours}
& 0.462 & 45.3\% & 0.491
& 2.84 & 0.038 & 0.0082 \\

\midrule

\multirow{2}{*}{QVHighlights~\cite{qvhighlight}}
& Fixed scheme
& 0.346 & 23.9\% & 0.395 & 9.87 & 0.107 & 0.0186 \\
& \textbf{Ours}
& 0.475 & 43.1\% & 0.528
& 4.92 & 0.054 & 0.0121 \\

\bottomrule
\end{tabular}
}
\end{table}

\subsection{Direct Analysis of Segmentation Quality}
\label{sec:seg_analysis}

To directly assess whether our video-specific segmentation mitigates semantic dilution, we conduct a post-hoc analysis using ground-truth (GT) moment annotations. These annotations are used only for evaluation and are never provided to TF-PRVR during retrieval. Let $G_q$ denote the GT moment for query $q$, and let $\mathcal{S}_q$ denote the generated segment candidates. We evaluate segmentation quality from five complementary perspectives. First, temporal alignment measures how well the generated hierarchy contains a segment aligned with the GT moment. We select the best-aligned segment
$
S_q^* = \arg\max_{S \in \mathcal{S}_q} \mathrm{tIoU}(S,G_q)
$
and report its mean tIoU and R@0.5. Second, GT coverage measures the fraction of the GT moment covered by $S_q^*$. Third, fragmentation counts the number of segments overlapping the GT moment at the scale containing $S_q^*$, where a lower value indicates that the relevant event is represented more compactly. Fourth, irrelevant-frame inclusion measures the fraction of each overlapping segment lying outside the GT moment. Finally, semantic coherence measures the intra-segment dispersion of frozen VLM features, where lower dispersion indicates more coherent segment content. As shown in Table~\ref{tab:seg_quality}, our video-specific segmentation consistently improves temporal alignment and GT coverage while reducing fragmentation, irrelevant-frame inclusion, and intra-segment feature dispersion across all evaluated datasets. These results provide direct evidence that the proposed segmentation produces better-aligned and more coherent temporal units than fixed decomposition, thereby mitigating semantic dilution.

\begin{table}[t]
\centering
\caption{
Budget-matched comparison of temporal boundary strategies.
Act, Cha, and QVH denote ActivityNet Captions~\cite{activitynet}, Charades-STA~\cite{charades-sta}, and QVHighlights~\cite{qvhighlight}, respectively.
All constrained variants use 160 candidates per video, while the unconstrained TF-PRVR uses a video-specific adaptive budget.
We report SumR on all datasets.
}
\vspace{0.4em}
\label{tab:budget_matched}
\footnotesize
\setlength{\tabcolsep}{10pt}

\resizebox{\linewidth}{!}{
\begin{tabular}{lccccc}
\toprule
Boundary Strategy & Budget 
& Act~\cite{activitynet}
& Cha~\cite{charades-sta}
& TVR~\cite{tvr}
& QVH~\cite{qvhighlight} \\
\midrule
Uniform fixed intervals (conventional)
& 160 & 191.8 & 80.1 & 175.4 & 206.1 \\

Adjacent feature dissimilarity
& 160 & 196.2 & 82.3 & 179.4 & 212.6 \\

Self-similarity matrix
& 160 & 197.0 & 83.8 & 180.1 & 213.9 \\

\textbf{Velocity-direction change (Ours)}
& 160 & 200.7 & 85.1 & 183.0 & 218.4 \\

\midrule

\textbf{Ours, unconstrained (video-specific)}
& Adaptive & 200.8 & 85.4 & 183.6 & 218.5 \\

\bottomrule
\end{tabular}
}
\end{table}

\subsection{Budget-Matched Boundary Comparison}
\label{sec:budget_matched}

To isolate the effect of boundary quality from the number of temporal candidates, we conduct a budget-matched comparison across different boundary construction strategies. Existing PRVR methods use 160 video representations per video, consisting of 128 frame-level and 32 clip-level representations. Accordingly, we restrict each compared segmentation variant to a fixed budget of 160 candidates per video. For each temporal scale, we rank internal boundary candidates by their response strength and retain them until the total budget is reached. When the detected local maxima are insufficient, we additionally select the highest-response positions under the same non-maximum-suppression constraint. In contrast, the unconstrained TF-PRVR adapts the number of segments to each video's temporal structure, averaging 164.8, 137.6, 145.3, and 152.7 candidates on ActivityNet Captions, Charades-STA, TVR, and QVHighlights, respectively. As shown in Table~\ref{tab:budget_matched}, the proposed velocity-direction signal consistently outperforms uniform fixed intervals, adjacent feature dissimilarity, and self-similarity under the same candidate budget. Its performance is also close to the unconstrained video-specific setting, indicating that the gain primarily comes from more informative boundary placement rather than a larger candidate set.

\begin{table}[t]
\centering
\caption{
Comparison with training-free baselines using the same frozen features.
Act, Cha, and QVH denote ActivityNet Captions~\cite{activitynet}, Charades-STA~\cite{charades-sta}, and QVHighlights~\cite{qvhighlight}, respectively.
All values are SumR.
}
\vspace{0.4em}
\label{tab:training_free_baseline}
\footnotesize
\setlength{\tabcolsep}{4pt}

\resizebox{\linewidth}{!}{
\begin{tabular}{lccc cccc}
\toprule
Method & Segmentation & GP & MAS
& Act & Cha & TVR & QVH \\
\midrule

Zero-shot
& -- & -- & --
& 180.4 & 77.5 & 172.2 & 197.6 \\

+ Single-scale mean pooling (32 clips)
& Fixed & -- & --
& 183.5 & 78.1 & 173.2 & 199.8 \\

+ Two-branch (clip + frame)
& Fixed & -- & --
& 186.6 & 78.8 & 174.4 & 202.1 \\

+ Multi-scale temporal pooling (3 scales)
& Fixed & -- & --
& 188.2 & 79.2 & 175.0 & 203.5 \\

+ Sliding-window multi-scale (exhaustive)
& Fixed & -- & --
& 189.5 & 79.5 & 175.3 & 204.8 \\

+ Graph propagation only
& Fixed & $\checkmark$ & --
& 190.6 & 80.0 & 175.8 & 205.9 \\

+ Graph propagation + moment-aware scoring
& Fixed & $\checkmark$ & $\checkmark$
& 191.8 & 80.5 & 176.3 & 207.0 \\

\midrule

+ Adaptive segmentation only
& Adaptive & -- & --
& 194.7 & 82.2 & 178.1 & 211.4 \\

+ Adaptive segmentation + GP
& Adaptive & $\checkmark$ & --
& 197.8 & 84.1 & 181.2 & 215.3 \\

\textbf{TF-PRVR (Full)}
& Adaptive & $\checkmark$ & $\checkmark$
& 200.8 & 85.4 & 183.6 & 218.5 \\

\bottomrule
\end{tabular}
}
\end{table}

\subsection{Comparison with Training-Free Baselines}
\label{sec:training_free_baselines}

To isolate the contribution of TF-PRVR from that of the frozen vision-language backbone, we conduct a controlled comparison with stronger training-free baselines using the same frozen features and no task-specific training. We progressively consider single-scale mean pooling, the conventional two-branch design, multi-scale temporal pooling, exhaustive sliding windows, graph propagation (GP), moment-aware scoring (MAS), and our adaptive hierarchical segmentation. As shown in Table~\ref{tab:training_free_baseline}, all fixed temporal aggregation variants improve over frame-level zero-shot retrieval. Among the fixed temporal aggregation baselines, exhaustive multi-scale sliding windows achieve the strongest performance. Replacing fixed segmentation with our adaptive hierarchy alone further improves SumR by 5.2, 2.7, 2.8, and 6.6 on ActivityNet Captions, Charades-STA, TVR, and QVHighlights, respectively. Under the same GP and MAS pipeline, adaptive segmentation improves over fixed segmentation by 9.0, 4.9, 7.3, and 11.5 SumR, while GP and MAS provide additional gains. These results show that the improvement of TF-PRVR cannot be attributed to the frozen backbone or stronger fixed temporal aggregation alone.

\begin{table}[t]
\centering
\caption{Ablation studies on ActivityNet Captions~\cite{activitynet}. We analyze the effects of DWT wavelet basis, the number of DWT decomposition levels, graph components, and propagation strength $\beta$.}
\vspace{-0.3em}
\label{tab:supple_ablation}
\tiny
\setlength{\tabcolsep}{3pt}

\begin{subtable}[t]{0.21\textwidth}
\centering
\caption{DWT wavelet basis.}
\vspace{-0.3em}
\label{tab:wavelet_basis_ablation}
{
\renewcommand{\arraystretch}{1.15}
\resizebox{\linewidth}{!}{
\begin{tabular}{lc}
\toprule
Variant & SumR \\
\midrule
Haar & 197.7 \\
db4 (default) & \textbf{200.8} \\
db8 & 200.6 \\
sym4 & 196.4 \\
\bottomrule
\end{tabular}
}
}
\end{subtable}%
\hfill
\begin{subtable}[t]{0.27\textwidth}
\centering
\caption{DWT levels.}
\vspace{-0.3em}
\label{tab:dwt_level_ablation}
{
\renewcommand{\arraystretch}{1.20}
\resizebox{\linewidth}{!}{
\begin{tabular}{lc}
\toprule
Variant & SumR \\
\midrule
$L=2$ & 195.6 \\
$L=3$ & 199.7 \\
$L=4$ & 200.3 \\
$L=\mathrm{auto}$ (default) & \textbf{200.8} \\
\bottomrule
\end{tabular}
}
}
\end{subtable}%
\hfill
\begin{subtable}[t]{0.27\textwidth}
\centering
\caption{Graph components.}
\vspace{-0.3em}
\label{tab:graph_component_ablation}
{
\renewcommand{\arraystretch}{1.16}
\resizebox{\linewidth}{!}{
\begin{tabular}{lc}
\toprule
Variant & SumR \\
\midrule
Visual intra-level only & 198.4 \\
Temporal intra-level only & 188.9 \\
Intra-level only & 195.2 \\
Inter-level only & 194.6 \\
Full graph (default) & \textbf{200.8} \\
\bottomrule
\end{tabular}
}
}
\end{subtable}%
\hfill
\begin{subtable}[t]{0.20\textwidth}
\centering
\caption{$\beta$ sensitivity.}
\vspace{-0.3em}
\label{tab:beta_sensitivity_ablation}
{
\renewcommand{\arraystretch}{1.0}
\resizebox{\linewidth}{!}{
\begin{tabular}{lc}
\toprule
$\beta$ & SumR \\
\midrule
0.1 & 197.4 \\
0.3 & 199.6 \\
0.5  & 200.4 \\
0.7 (default) & \textbf{200.8} \\
0.9 & 198.6 \\
\bottomrule
\end{tabular}
}
}
\end{subtable}

\end{table}

\subsection{DWT Wavelet Basis}
We examine the effect of the wavelet basis on ActivityNet Captions. As shown in Table~\ref{tab:wavelet_basis_ablation}, db4 achieves the highest SumR of 200.8, closely followed by db8 at 200.6, while Haar and sym4 obtain 197.7 and 196.4, respectively. These results suggest that the wavelet basis can affect retrieval performance within the same decomposition framework. We use db4 for this setting and select the wavelet basis separately for each dataset.

\subsection{Number of DWT Decomposition Levels}
We evaluate the effect of the number of DWT decomposition levels. As shown in Table~\ref{tab:supple_ablation} (b), using a small number of levels, such as $L=2$, yields a lower SumR of 195.6, indicating insufficient temporal granularity. Increasing the number of levels improves performance, with $L=3$ and $L=4$ achieving 199.7 and 200.3, respectively. The automatic level selection achieves the best SumR of 200.8, showing that adapting the decomposition depth to each video better captures video-specific temporal structures. We therefore adopt automatic DWT level selection as the default setting.

\subsection{Graph Components}
We examine the contribution of each graph component. As shown in Table~\ref{tab:supple_ablation} (c), using only visual intra-level edges achieves a SumR of 198.4, while using only temporal intra-level edges drops to 188.9. This indicates that semantic similarity is more informative than temporal proximity alone, but temporal structure still provides complementary cues. Intra-level-only and inter-level-only variants obtain 195.2 and 194.6, respectively, showing that both within-scale and cross-scale connections are necessary. The full graph achieves the best SumR of 200.8 by combining visual, temporal, and hierarchical relationships, validating the design of our unified multi-scale graph.

\subsection{Propagation Strength $\beta$}
We analyze the sensitivity to the propagation coefficient $\beta$, which balances the initial query-segment relevance and the propagated graph signal. As shown in Table~\ref{tab:supple_ablation} (d), small values such as $\beta=0.1$ and $\beta=0.3$ underutilize graph propagation, achieving SumR scores of 197.4 and 199.6. Performance improves as $\beta$ increases, reaching the best SumR of 200.8 at $\beta=0.7$. However, a larger value of $\beta=0.9$ reduces SumR to 198.6, suggesting that excessive propagation can oversmooth query-specific relevance.

\begin{table}[t]
\centering
\caption{SumR performance comparison across video length groups on ActivityNet Captions~\cite{activitynet}, Charades-STA~\cite{charades-sta}, and TVR~\cite{tvr}, using CLIP-L/14~\cite{CLIP} features. Videos in each benchmark are partitioned into full, short, and long groups according to video duration.}
\label{tab:supple_video_length}
\small
\setlength{\tabcolsep}{8pt}

\resizebox{\linewidth}{!}{
\begin{tabular}{lccc ccc ccc}
\toprule
\multirow{2}{*}{Method}
& \multicolumn{3}{c}{ActivityNet Captions~\cite{activitynet}}
& \multicolumn{3}{c}{Charades-STA~\cite{charades-sta}}
& \multicolumn{3}{c}{TVR~\cite{tvr}} \\
\cmidrule(lr){2-4} \cmidrule(lr){5-7} \cmidrule(lr){8-10}
& Full & Short & Long
& Full & Short & Long
& Full & Short & Long \\
\midrule
TF-PRVR (Ours)
& 193.9 & 193.7 & 194.0
& 80.8 & 80.7 & 80.8
& 178.3 & 178.4 & 178.1 \\
\bottomrule
\end{tabular}
}

\end{table}
\subsection{Performance Comparison Across Video Length}
We analyze the robustness of TF-PRVR across different video lengths. As discussed in Table~3 of the main paper, we show that existing PRVR methods exhibit different performance trends on short and long videos depending on the dataset-specific moment duration. This is mainly because fixed clip decomposition uses the same number of clips for all videos, causing the temporal span of each clip to vary with video length. In contrast, TF-PRVR constructs video-specific hierarchical segments from semantic transition boundaries rather than fixed intervals. As shown in Table~\ref{tab:supple_video_length}, TF-PRVR maintains highly stable performance across full, short, and long groups on all three benchmarks. On ActivityNet Captions, TF-PRVR achieves 193.7 on short videos and 194.0 on long videos, showing almost no performance gap. Similar trends are observed on Charades-STA, where the scores remain 80.7 and 80.8, and on TVR, where TF-PRVR obtains 178.4 and 178.1 for short and long videos, respectively. These results indicate that TF-PRVR is less sensitive to video duration and can consistently handle diverse temporal structures by adapting its segmentation to each video.

\subsection{Failure Case}
While our DWT-based decomposition generally captures semantic transitions, low-variance temporal signals can lead to suboptimal boundaries. As shown in Fig.~\ref{fig:vis}, the recurring indoor background (Frames \#2, \#16, \#30, and \#85) produces relatively flat responses, causing coarser levels to occasionally merge semantically distinct moments. However, finer-scale segmentation provides more granular alternatives, while multi-scale graph consistency further mitigates errors from any single level.




\begin{table}[t]
\centering
\caption{
Deployment-level efficiency comparison on ActivityNet Captions~\cite{activitynet}.
End-to-end time includes sampled-frame decoding, frozen-feature extraction, index construction, and one text query.
}
\vspace{0.4em}
\label{tab:deployment_efficiency}
\footnotesize
\setlength{\tabcolsep}{12pt}

\resizebox{\linewidth}{!}{
\begin{tabular}{lcccc}
\toprule
Method
& \shortstack{Indexed Query\\Latency (ms)}
& \shortstack{End-to-End\\Time (s)}
& \shortstack{Indexed Query\\Memory (GB)}
& \shortstack{Peak Deployment\\Memory (GB)} \\
\midrule

MS-SL~\cite{ms-sl}
& 54.33 & 48.74 & 7.57 & 8.29 \\

GMMFormerv2~\cite{gmmformerv2}
& 9.74 & 48.31 & 1.94 & 2.66 \\

MSC-PRVR~\cite{msc-prvr}
& 9.78 & 48.36 & 0.48 & 1.20 \\

\textbf{TF-PRVR (Ours)}
& 17.24 & 49.08 & 3.80 & 4.52 \\

\bottomrule
\end{tabular}
}
\end{table}
\subsection{Computational Efficiency}

We measure deployment-level efficiency on ActivityNet Captions with 4,430 videos. The end-to-end measurement includes sampled-frame decoding, frozen-feature extraction, index construction, and one text query under identical hardware and batch settings. As shown in Table~\ref{tab:deployment_efficiency}, TF-PRVR incurs higher indexed query latency and memory than compact training-based methods. Compared with MSC-PRVR, latency increases from 9.78 ms to 17.24 ms and indexed memory from 0.48 GB to 3.80 GB. However, the end-to-end time increases only from 48.36 s to 49.08 s, a 0.72 s (1.5\%) difference. Because hierarchy, graph, and propagation-kernel construction are query-independent, these costs are amortized across subsequent queries. Thus, the main deployment trade-off is higher index and query-time memory rather than prohibitive latency.

\renewcommand{\thesection}{C}
\section{Comparison with Video Moment Retrieval and Temporal Grounding}
PRVR is closely related to video moment retrieval (VMR)~\cite{gao2021fast, luo2023towards, zhang2019man, anne2017localizing, liu2021context}, also known as temporal sentence grounding, and video corpus moment retrieval (VCMR)~\cite{tvr, li2020hero, zhang2021video, jung2022modal}. VMR and temporal grounding typically localize a query-relevant moment within a given video, while VCMR jointly retrieves a relevant video and its temporal moment from a corpus. In contrast, PRVR ranks untrimmed videos according to whether they contain a relevant moment, without requiring moment-level supervision or temporal boundaries as output. Recent zero-shot VMR methods leverage frozen vision-language models for training-free temporal localization~\cite{luo2024zero, wang2022prompt}, while prior grounding methods employ multi-scale temporal candidates or graph-based reasoning~\cite{zhang2019man, soldan2021vlg}. Since existing VMR, and VCMR methods are optimized for precise temporal boundary prediction, their architectures are not naturally aligned with the PRVR protocol. PRVR instead requires deriving a reliable video-level relevance score over a corpus without relying on moment-level temporal boundary supervision. Consequently, while we share foundational cross-modal concepts, directly adapting these localization-centric models as baselines for PRVR is conceptually misaligned and non-trivial without substantial structural modifications. Furthermore, since acquiring precise moment-level boundary annotations for massive video corpora is highly labor-intensive, PRVR offers a more scalable and practical solution for real-world video search systems.

\renewcommand{\thesection}{D}
\section{Positive and Negative Societal Impacts}

\noindent\textbf{Positive Societal Impacts.}
TF-PRVR aims to improve access to relevant information in large-scale untrimmed video collections. By retrieving videos that contain query-relevant moments without requiring task-specific training or moment-level annotations, our framework can reduce the cost of building video retrieval systems and make them more adaptable to diverse real-world domains. This may benefit applications such as educational video search, assistive media browsing, archival retrieval, and efficient content discovery, where users often seek specific information within long videos. The training-free nature of TF-PRVR also reduces the need for large task-specific labeled datasets, potentially lowering computational and annotation costs.

\noindent\textbf{Negative Societal Impacts.}
As with other video retrieval systems, TF-PRVR could be misused for privacy-invasive search or surveillance. It may also inherit biases and failure modes from the underlying vision-language backbone, leading to uneven retrieval quality across domains or demographic groups. Practical deployment should therefore consider privacy protection, access control, dataset governance, and careful evaluation of backbone biases.

\clearpage
\section*{NeurIPS Paper Checklist}

\begin{enumerate}

\item {\bf Claims}
    \item[] Question: Do the main claims made in the abstract and introduction accurately reflect the paper's contributions and scope?
    \item[] Answer: \answerYes{} 
    \item[] Justification: The abstract and introduction state the contributions of the paper.
    \item[] Guidelines:
    \begin{itemize}
        \item The answer \answerNA{} means that the abstract and introduction do not include the claims made in the paper.
        \item The abstract and/or introduction should clearly state the claims made, including the contributions made in the paper and important assumptions and limitations. A \answerNo{} or \answerNA{} answer to this question will not be perceived well by the reviewers. 
        \item The claims made should match theoretical and experimental results, and reflect how much the results can be expected to generalize to other settings. 
        \item It is fine to include aspirational goals as motivation as long as it is clear that these goals are not attained by the paper. 
    \end{itemize}

\item {\bf Limitations}
    \item[] Question: Does the paper discuss the limitations of the work performed by the authors?
    \item[] Answer: \answerYes{} 
    \item[] Justification: The limitation section is provided in the main paper.
    \item[] Guidelines:
    \begin{itemize}
        \item The answer \answerNA{} means that the paper has no limitation while the answer \answerNo{} means that the paper has limitations, but those are not discussed in the paper. 
        \item The authors are encouraged to create a separate ``Limitations'' section in their paper.
        \item The paper should point out any strong assumptions and how robust the results are to violations of these assumptions (e.g., independence assumptions, noiseless settings, model well-specification, asymptotic approximations only holding locally). The authors should reflect on how these assumptions might be violated in practice and what the implications would be.
        \item The authors should reflect on the scope of the claims made, e.g., if the approach was only tested on a few datasets or with a few runs. In general, empirical results often depend on implicit assumptions, which should be articulated.
        \item The authors should reflect on the factors that influence the performance of the approach. For example, a facial recognition algorithm may perform poorly when image resolution is low or images are taken in low lighting. Or a speech-to-text system might not be used reliably to provide closed captions for online lectures because it fails to handle technical jargon.
        \item The authors should discuss the computational efficiency of the proposed algorithms and how they scale with dataset size.
        \item If applicable, the authors should discuss possible limitations of their approach to address problems of privacy and fairness.
        \item While the authors might fear that complete honesty about limitations might be used by reviewers as grounds for rejection, a worse outcome might be that reviewers discover limitations that aren't acknowledged in the paper. The authors should use their best judgment and recognize that individual actions in favor of transparency play an important role in developing norms that preserve the integrity of the community. Reviewers will be specifically instructed to not penalize honesty concerning limitations.
    \end{itemize}

\item {\bf Theory assumptions and proofs}
    \item[] Question: For each theoretical result, does the paper provide the full set of assumptions and a complete (and correct) proof?
    \item[] Answer: \answerNA{} 
    \item[] Justification: This paper does not include formal theorems or proofs.
    \item[] Guidelines:
    \begin{itemize}
        \item The answer \answerNA{} means that the paper does not include theoretical results. 
        \item All the theorems, formulas, and proofs in the paper should be numbered and cross-referenced.
        \item All assumptions should be clearly stated or referenced in the statement of any theorems.
        \item The proofs can either appear in the main paper or the supplemental material, but if they appear in the supplemental material, the authors are encouraged to provide a short proof sketch to provide intuition. 
        \item Inversely, any informal proof provided in the core of the paper should be complemented by formal proofs provided in appendix or supplemental material.
        \item Theorems and Lemmas that the proof relies upon should be properly referenced. 
    \end{itemize}

    \item {\bf Experimental result reproducibility}
    \item[] Question: Does the paper fully disclose all the information needed to reproduce the main experimental results of the paper to the extent that it affects the main claims and/or conclusions of the paper (regardless of whether the code and data are provided or not)?
    \item[] Answer: \answerYes{} 
    \item[] Justification: We discuss the details required to reproduce the experiments in this paper in the Experiments section.
    \item[] Guidelines:
    \begin{itemize}
        \item The answer \answerNA{} means that the paper does not include experiments.
        \item If the paper includes experiments, a \answerNo{} answer to this question will not be perceived well by the reviewers: Making the paper reproducible is important, regardless of whether the code and data are provided or not.
        \item If the contribution is a dataset and\slash or model, the authors should describe the steps taken to make their results reproducible or verifiable. 
        \item Depending on the contribution, reproducibility can be accomplished in various ways. For example, if the contribution is a novel architecture, describing the architecture fully might suffice, or if the contribution is a specific model and empirical evaluation, it may be necessary to either make it possible for others to replicate the model with the same dataset, or provide access to the model. In general. releasing code and data is often one good way to accomplish this, but reproducibility can also be provided via detailed instructions for how to replicate the results, access to a hosted model (e.g., in the case of a large language model), releasing of a model checkpoint, or other means that are appropriate to the research performed.
        \item While NeurIPS does not require releasing code, the conference does require all submissions to provide some reasonable avenue for reproducibility, which may depend on the nature of the contribution. For example
        \begin{enumerate}
            \item If the contribution is primarily a new algorithm, the paper should make it clear how to reproduce that algorithm.
            \item If the contribution is primarily a new model architecture, the paper should describe the architecture clearly and fully.
            \item If the contribution is a new model (e.g., a large language model), then there should either be a way to access this model for reproducing the results or a way to reproduce the model (e.g., with an open-source dataset or instructions for how to construct the dataset).
            \item We recognize that reproducibility may be tricky in some cases, in which case authors are welcome to describe the particular way they provide for reproducibility. In the case of closed-source models, it may be that access to the model is limited in some way (e.g., to registered users), but it should be possible for other researchers to have some path to reproducing or verifying the results.
        \end{enumerate}
    \end{itemize}

\item {\bf Open access to data and code}
    \item[] Question: Does the paper provide open access to the data and code, with sufficient instructions to faithfully reproduce the main experimental results, as described in supplemental material?
    \item[] Answer: \answerYes{} 
    \item[] Justification: We provide the source code.
    \item[] Guidelines:
    \begin{itemize}
        \item The answer \answerNA{} means that paper does not include experiments requiring code.
        \item Please see the NeurIPS code and data submission guidelines (\url{https://neurips.cc/public/guides/CodeSubmissionPolicy}) for more details.
        \item While we encourage the release of code and data, we understand that this might not be possible, so \answerNo{} is an acceptable answer. Papers cannot be rejected simply for not including code, unless this is central to the contribution (e.g., for a new open-source benchmark).
        \item The instructions should contain the exact command and environment needed to run to reproduce the results. See the NeurIPS code and data submission guidelines (\url{https://neurips.cc/public/guides/CodeSubmissionPolicy}) for more details.
        \item The authors should provide instructions on data access and preparation, including how to access the raw data, preprocessed data, intermediate data, and generated data, etc.
        \item The authors should provide scripts to reproduce all experimental results for the new proposed method and baselines. If only a subset of experiments are reproducible, they should state which ones are omitted from the script and why.
        \item At submission time, to preserve anonymity, the authors should release anonymized versions (if applicable).
        \item Providing as much information as possible in supplemental material (appended to the paper) is recommended, but including URLs to data and code is permitted.
    \end{itemize}

\item {\bf Experimental setting/details}
    \item[] Question: Does the paper specify all the training and test details (e.g., data splits, hyperparameters, how they were chosen, type of optimizer) necessary to understand the results?
    \item[] Answer: \answerYes{} 
    \item[] Justification: Since our framework is training-free, there are no training details such as optimizers or learning rates. All relevant details and hyperparameters are specified in the Experiments section and the supplementary material. Standard dataset splits are used for all benchmarks.
    \item[] Guidelines:
    \begin{itemize}
        \item The answer \answerNA{} means that the paper does not include experiments.
        \item The experimental setting should be presented in the core of the paper to a level of detail that is necessary to appreciate the results and make sense of them.
        \item The full details can be provided either with the code, in appendix, or as supplemental material.
    \end{itemize}

\item {\bf Experiment statistical significance}
    \item[] Question: Does the paper report error bars suitably and correctly defined or other appropriate information about the statistical significance of the experiments?
    \item[] Answer: \answerNo{} 
    \item[] Justification:  We report standard retrieval metrics on fixed benchmark splits, following the evaluation protocols of prior PRVR work.
    \item[] Guidelines:
    \begin{itemize}
        \item The answer \answerNA{} means that the paper does not include experiments.
        \item The authors should answer \answerYes{} if the results are accompanied by error bars, confidence intervals, or statistical significance tests, at least for the experiments that support the main claims of the paper.
        \item The factors of variability that the error bars are capturing should be clearly stated (for example, train/test split, initialization, random drawing of some parameter, or overall run with given experimental conditions).
        \item The method for calculating the error bars should be explained (closed form formula, call to a library function, bootstrap, etc.)
        \item The assumptions made should be given (e.g., Normally distributed errors).
        \item It should be clear whether the error bar is the standard deviation or the standard error of the mean.
        \item It is OK to report 1-sigma error bars, but one should state it. The authors should preferably report a 2-sigma error bar than state that they have a 96\% CI, if the hypothesis of Normality of errors is not verified.
        \item For asymmetric distributions, the authors should be careful not to show in tables or figures symmetric error bars that would yield results that are out of range (e.g., negative error rates).
        \item If error bars are reported in tables or plots, the authors should explain in the text how they were calculated and reference the corresponding figures or tables in the text.
    \end{itemize}

\item {\bf Experiments compute resources}
    \item[] Question: For each experiment, does the paper provide sufficient information on the computer resources (type of compute workers, memory, time of execution) needed to reproduce the experiments?
    \item[] Answer: \answerYes{} 
    \item[] Justification: The paper provides the compute resources used for the reported experiments, including the GPU/CPU environment.
    \item[] Guidelines:
    \begin{itemize}
        \item The answer \answerNA{} means that the paper does not include experiments.
        \item The paper should indicate the type of compute workers CPU or GPU, internal cluster, or cloud provider, including relevant memory and storage.
        \item The paper should provide the amount of compute required for each of the individual experimental runs as well as estimate the total compute. 
        \item The paper should disclose whether the full research project required more compute than the experiments reported in the paper (e.g., preliminary or failed experiments that didn't make it into the paper). 
    \end{itemize}
    
\item {\bf Code of ethics}
    \item[] Question: Does the research conducted in the paper conform, in every respect, with the NeurIPS Code of Ethics \url{https://neurips.cc/public/EthicsGuidelines}?
    \item[] Answer: \answerYes{} 
    \item[] Justification: We have reviewed the NeurIPS Code of Ethics and confirm that our research conforms with it in every respect. We discuss potential privacy, fairness, and societal concerns in the supplementary material.
    \item[] Guidelines:
    \begin{itemize}
        \item The answer \answerNA{} means that the authors have not reviewed the NeurIPS Code of Ethics.
        \item If the authors answer \answerNo, they should explain the special circumstances that require a deviation from the Code of Ethics.
        \item The authors should make sure to preserve anonymity (e.g., if there is a special consideration due to laws or regulations in their jurisdiction).
    \end{itemize}

\item {\bf Broader impacts}
    \item[] Question: Does the paper discuss both potential positive societal impacts and negative societal impacts of the work performed?
    \item[] Answer: \answerYes{} 
    \item[] Justification: Societal impacts are included in the supplementary material.
    \item[] Guidelines:
    \begin{itemize}
        \item The answer \answerNA{} means that there is no societal impact of the work performed.
        \item If the authors answer \answerNA{} or \answerNo, they should explain why their work has no societal impact or why the paper does not address societal impact.
        \item Examples of negative societal impacts include potential malicious or unintended uses (e.g., disinformation, generating fake profiles, surveillance), fairness considerations (e.g., deployment of technologies that could make decisions that unfairly impact specific groups), privacy considerations, and security considerations.
        \item The conference expects that many papers will be foundational research and not tied to particular applications, let alone deployments. However, if there is a direct path to any negative applications, the authors should point it out. For example, it is legitimate to point out that an improvement in the quality of generative models could be used to generate Deepfakes for disinformation. On the other hand, it is not needed to point out that a generic algorithm for optimizing neural networks could enable people to train models that generate Deepfakes faster.
        \item The authors should consider possible harms that could arise when the technology is being used as intended and functioning correctly, harms that could arise when the technology is being used as intended but gives incorrect results, and harms following from (intentional or unintentional) misuse of the technology.
        \item If there are negative societal impacts, the authors could also discuss possible mitigation strategies (e.g., gated release of models, providing defenses in addition to attacks, mechanisms for monitoring misuse, mechanisms to monitor how a system learns from feedback over time, improving the efficiency and accessibility of ML).
    \end{itemize}
    
\item {\bf Safeguards}
    \item[] Question: Does the paper describe safeguards that have been put in place for responsible release of data or models that have a high risk for misuse (e.g., pre-trained language models, image generators, or scraped datasets)?
    \item[] Answer: \answerNA{} 
    \item[] Justification: This work does not release a new high-risk model, pre-trained generative model, scraped dataset, or data containing newly collected sensitive information.
    \item[] Guidelines:
    \begin{itemize}
        \item The answer \answerNA{} means that the paper poses no such risks.
        \item Released models that have a high risk for misuse or dual-use should be released with necessary safeguards to allow for controlled use of the model, for example by requiring that users adhere to usage guidelines or restrictions to access the model or implementing safety filters. 
        \item Datasets that have been scraped from the Internet could pose safety risks. The authors should describe how they avoided releasing unsafe images.
        \item We recognize that providing effective safeguards is challenging, and many papers do not require this, but we encourage authors to take this into account and make a best faith effort.
    \end{itemize}

\item {\bf Licenses for existing assets}
    \item[] Question: Are the creators or original owners of assets (e.g., code, data, models), used in the paper, properly credited and are the license and terms of use explicitly mentioned and properly respected?
    \item[] Answer: \answerYes{} 
    \item[] Justification: All datasets and pre-trained models used in this paper are properly cited in the references.
    \item[] Guidelines:
    \begin{itemize}
        \item The answer \answerNA{} means that the paper does not use existing assets.
        \item The authors should cite the original paper that produced the code package or dataset.
        \item The authors should state which version of the asset is used and, if possible, include a URL.
        \item The name of the license (e.g., CC-BY 4.0) should be included for each asset.
        \item For scraped data from a particular source (e.g., website), the copyright and terms of service of that source should be provided.
        \item If assets are released, the license, copyright information, and terms of use in the package should be provided. For popular datasets, \url{paperswithcode.com/datasets} has curated licenses for some datasets. Their licensing guide can help determine the license of a dataset.
        \item For existing datasets that are re-packaged, both the original license and the license of the derived asset (if it has changed) should be provided.
        \item If this information is not available online, the authors are encouraged to reach out to the asset's creators.
    \end{itemize}

\item {\bf New assets}
    \item[] Question: Are new assets introduced in the paper well documented and is the documentation provided alongside the assets?
    \item[] Answer: \answerNA{} 
    \item[] Justification: This paper does not release new datasets, models, or other assets.
    \item[] Guidelines:
    \begin{itemize}
        \item The answer \answerNA{} means that the paper does not release new assets.
        \item Researchers should communicate the details of the dataset\slash code\slash model as part of their submissions via structured templates. This includes details about training, license, limitations, etc. 
        \item The paper should discuss whether and how consent was obtained from people whose asset is used.
        \item At submission time, remember to anonymize your assets (if applicable). You can either create an anonymized URL or include an anonymized zip file.
    \end{itemize}

\item {\bf Crowdsourcing and research with human subjects}
    \item[] Question: For crowdsourcing experiments and research with human subjects, does the paper include the full text of instructions given to participants and screenshots, if applicable, as well as details about compensation (if any)? 
    \item[] Answer: \answerNA{} 
    \item[] Justification: This paper does not involve crowdsourcing or research with human subjects. All experiments are conducted on existing publicly available datasets.
    \item[] Guidelines:
    \begin{itemize}
        \item The answer \answerNA{} means that the paper does not involve crowdsourcing nor research with human subjects.
        \item Including this information in the supplemental material is fine, but if the main contribution of the paper involves human subjects, then as much detail as possible should be included in the main paper. 
        \item According to the NeurIPS Code of Ethics, workers involved in data collection, curation, or other labor should be paid at least the minimum wage in the country of the data collector. 
    \end{itemize}

\item {\bf Institutional review board (IRB) approvals or equivalent for research with human subjects}
    \item[] Question: Does the paper describe potential risks incurred by study participants, whether such risks were disclosed to the subjects, and whether Institutional Review Board (IRB) approvals (or an equivalent approval/review based on the requirements of your country or institution) were obtained?
    \item[] Answer: \answerNA{} 
    \item[] Justification: \answerNA{}
    \item[] Guidelines:
    \begin{itemize}
        \item The answer \answerNA{} means that the paper does not involve crowdsourcing nor research with human subjects.
        \item Depending on the country in which research is conducted, IRB approval (or equivalent) may be required for any human subjects research. If you obtained IRB approval, you should clearly state this in the paper. 
        \item We recognize that the procedures for this may vary significantly between institutions and locations, and we expect authors to adhere to the NeurIPS Code of Ethics and the guidelines for their institution. 
        \item For initial submissions, do not include any information that would break anonymity (if applicable), such as the institution conducting the review.
    \end{itemize}

\item {\bf Declaration of LLM usage}
    \item[] Question: Does the paper describe the usage of LLMs if it is an important, original, or non-standard component of the core methods in this research? Note that if the LLM is used only for writing, editing, or formatting purposes and does \emph{not} impact the core methodology, scientific rigor, or originality of the research, declaration is not required.
    \item[] Answer:  \answerNA{} 
    \item[] Justification: LLMs were used solely for grammar checking and editing.
    \item[] Guidelines:
    \begin{itemize}
        \item The answer \answerNA{} means that the core method development in this research does not involve LLMs as any important, original, or non-standard components.
        \item Please refer to our LLM policy in the NeurIPS handbook for what should or should not be described.
    \end{itemize}

\end{enumerate}

\end{document}